\documentclass{new_tlp}

\newcommand\bcmdtab{\noindent\bgroup\tabcolsep=0pt
  \begin{tabular}{@{}p{10pc}@{}p{20pc}@{}}}
\newcommand\ecmdtab{\end{tabular}\egroup}

\usepackage{times}

\usepackage{color}
\usepackage{xspace}
\usepackage{graphicx} 
\usepackage{amssymb} 
\usepackage[shortlabels]{enumitem} 
\usepackage{multirow,caption,hhline} 
\usepackage{balance}
\usepackage{algorithm}
\usepackage[noend]{algorithmic}
\newcommand{\nop}[1]{}
\newtheorem{definition}{Definition}
\newtheorem{example}{Example}
\newtheorem{proposition}{Proposition}
\newtheorem{theorem}{Theorem}

\usepackage{mathtools}

\newcommand{\supportplus}{\ensuremath{\overset{+}{\Rightarrow}}}

\newcommand{\wrt}{w.r.t.\xspace}

\renewcommand{\iff}{iff\xspace}

\newcommand{\Set}[1]{\ensuremath{\mathbf{#1}}}

\newcommand{\AFargsNew}{\ensuremath{A}\xspace}
\newcommand{\AFattRelNew}{\ensuremath{\Omega}\xspace}
\newcommand{\AFNew}{\ensuremath{\langle \AFargsNew, \AFattRelNew\rangle}\xspace}

\def\aaf{{\mathcal{A}}}

\def\I{{\mathcal{I}}}

\def\reach0{\it Reach\mbox{$\!\!_{{_\aaf}_0}\!$}}

\def\+{\mbox{+}}
\def\-{\mbox{-}}
\def\<{\langle}
\def\>{\rangle}

\newcommand{\ldot}{\,{\bf .}\,}
\newcommand{\Def}{\textsc{Def}}
\newcommand{\Acc}{\textsc{Acc}}
\def\={\mbox{=}}

\def\true{\mbox{$\tt true$}}
\def\false{\mbox{$\tt false$}}
\def\undec{\mbox{$\tt undec$}}
\def\a{\mbox{$\tt a$}}
\def\b{\mbox{$\tt b$}}
\def\c{\mbox{$\tt c$}}
\def\d{\mbox{$\tt d$}}
\def\Sb{\Set{S}}
\def\sb{\Set{s}}
\def\tb{\Set{t}}

\newcommand{\blue}[1]{{\color{black}{#1}}}

\newcommand{\imply}{\Rightarrow}
\newcommand{\Deltan}{\Delta_n}
\newcommand{\Deltad}{\Delta_d}

\newcommand{\reaf}{Rec-AF}

  \title[On the Semantics of Argumentation Frameworks:  A Logic Programming Approach]
        {On the Semantics of Abstract Argumentation Frameworks:  A Logic Programming Approach}

  \author[G. Alfano, S. Greco, F. Parisi, and I. Trubitsyna]
         {Gianvincenzo Alfano, Sergio Greco, Francesco Parisi, and Irina Trubitsyna\\
         DIMES Department, University of Calabria, Rende, Italy\\
         \email{$\{$g.alfano,greco,fparisi,i.trubitsyna$\}$@dimes.unical.it}}

\submitted{}
\revised{}
\accepted{}
\pagerange{\pageref{firstpage}--\pageref{lastpage}}
\doi{S1471068401001193}

\newtheorem{lemma}{Lemma}[section]

\begin{document}

\label{firstpage}

\maketitle

  \begin{abstract}
Recently there has been an increasing interest in frameworks extending Dung's abstract Argumentation Framework (AF).
Popular extensions include bipolar AFs and AFs with recursive attacks 
and necessary supports.
Although the relationships between AF semantics and Partial Stable Models (PSMs) 
of logic programs has been deeply investigated, 
this is not the case for more general frameworks 
extending AF. 

In this paper we explore the relationships between AF-based frameworks and PSMs. 
We show that every AF-based framework $\Delta$ can be translated into a 
logic program $P_\Delta$ so that the extensions prescribed by different semantics 
of $\Delta$ coincide with subsets of the PSMs of $P_\Delta$.
We provide a logic programming approach that characterizes, 
in an elegant and uniform way, the semantics of several AF-based frameworks.
This result allows also to define the semantics for new AF-based frameworks, 
such as AFs with recursive attacks and recursive deductive supports. 

Under consideration for publication in Theory and Practice of Logic Programming.

  \end{abstract}

  \begin{keywords}
    abstract argumentation, argumentation semantics, partial stable models
  \end{keywords}

\section{Introduction}\label{Sec_Intro}

Formal argumentation has emerged as one of the important fields in
Artificial Intelligence~\cite{BenchCapon2007619,Rahwan-Simari-Book}.
In particular, Dung's abstract Argumentation Framework (AF) is a simple, yet powerful formalism for modelling disputes between two or more agents~\cite{Dung95}.
An AF consists of a set of \emph{arguments} and a binary \emph{attack} relation over the set of arguments that specifies the \textit{interactions} between arguments:
intuitively, if argument $a$ attacks argument $b$, then $b$ is acceptable only if $a$ is not.
Hence, arguments are abstract entities whose role is entirely determined by the interactions specified by the attack relation.

Dung's framework has been extended in many different ways, including the introduction of new kinds of interactions between arguments and/or attacks.
In particular, the class of Bipolar Argumentation Frameworks (BAFs) is an interesting extension of the AF which allows for also modelling the \textit{support} between arguments~\cite{NouiouaR11,VillataBGT12}.
Further extensions consider second-order interactions~\cite{VillataBGT12}, e.g., attacks to attacks/supports, as well as more general forms of interactions such as recursive AFs where attacks can be recursively attacked~\cite{Baroni-Cerutti-Giacomin-Guida11,CayrolFCL17} and  recursive BAFs, where attacks/supports can be recursively attacked/supported~\cite{Gottifredi-Cohen-Garcia-Simari18,CayrolFCL18}. 
An overview of the extensions of the Dung's framework is provided at the end of this section.

\begin{figure}[t!]
\begin{minipage}[b]{.4\textwidth}
\centering
\includegraphics[scale=0.5]{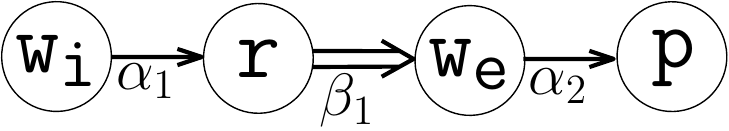}
\caption{\small BAF of Example~\ref{ex1:intro}}\label{fig:asaf-one}
\end{minipage}
\hspace*{+20mm}
\begin{minipage}[b]{.4\textwidth}
\centering
\includegraphics[scale=0.5]{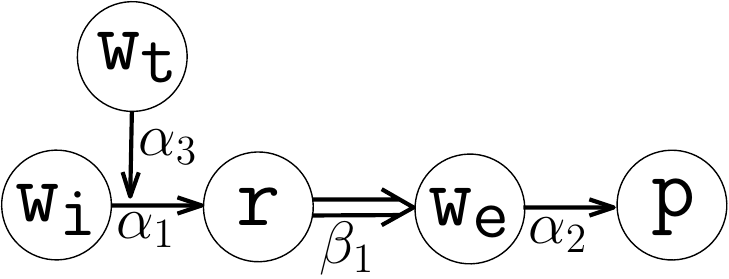}
\caption{\small Recursive BAF of Example~\ref{ex1:intro}}
\label{fig:asaf-two}
\end{minipage}
\end{figure}

\begin{example}\label{ex1:intro}\rm 
Consider a scenario for deciding whether to play tennis.
Assume we have the following arguments: 
$\tt w_i$ (it is windy), 
$\tt r$ (it is raining), 
$\tt w_e$ (the court is wet), 
$\tt p$ (play tennis), 
and the logical implications: 
($\alpha_1$) if it is windy, then it does not rain, 
($\alpha_2$) if the court is wet, then we do not play tennis, and 
($\beta_1$) if it is raining, then the court is wet. 
This situation can be modelled using the BAF shown in Figure~\ref{fig:asaf-one}, where
the implications $\alpha_1$ and $\alpha_2$ are \emph{attacks} (denoted by $\rightarrow$), and the implication $\beta_1$ is a \emph{support} (denoted by $\Rightarrow$). 

Now assume that there also exists an argument $\tt w_t$ (we are in the winter season) that attacks  the implication $\alpha_1$ (in the winter season, implication $\alpha_1$ cannot be applied). The new scenario can be modeled by the recursive BAF shown in Figure~\ref{fig:asaf-two} where the new attack is named as $\alpha_3$.~\hfill~$\square$
\end{example}

Several interpretations of the notion of support have been proposed~\cite{CayrolL13,CohenGGS14}.
Intuitively, the way the support is interpreted changes the set of \emph{extensions} (i.e., the set of acceptable elements) of an argumentation framework.
For instance, the (unique complete) extension of the BAF shown in Figure~\ref{fig:asaf-one} is the set $\{\tt w_i, p\}$ under the so-called \textit{necessary} interpretation of support, while it is $\{\tt w_i, w_e\}$ under the \textit{deductive} interpretation. 

Following Dung's approach, the meaning of recursive AF-based frameworks is still given by relying on the concept of extension.
However, the extensions of an \textit{AF with Recursive Attacks (AFRA)}~\cite{Baroni-Cerutti-Giacomin-Guida11} and of an \textit{Attack-Support Argumentation Framework (ASAF)}~\cite{CohenGGS15,Gottifredi-Cohen-Garcia-Simari18} also include the (names of) attacks and supports that intuitively contribute to determine the set of accepted arguments. Particularly, the acceptability of an attack is related to the acceptability of its source argument: 
an attack in the AFRA is defeated even when its source argument is defeated.
This is not the case for \textit{Recursive AF (RAF)}~\cite{CayrolFCL17}
and \textit{Recursive AF with Necessities (RAFN)} frameworks~\cite{CayrolFCL18}, which offer a different semantics for recursive AFs and recursive BAFs with necessary supports, respectively.

Recently there has been an increasing interest in studying the relationships between argumentation frameworks and logic programming (LP).
\blue{
In particular, 
the semantic equivalence between complete extensions in AF and
3-valued stable models in LP was first established in~\cite{WuCG09}.
Then,}
the relationships of LP with AF have been \blue{further} studied in~\cite{CaminadaSAD15}, 
\blue{whereas those with Assumption-Based Argumentation~\cite{BondarenkoDKT97,CravenT16} have been considered in~\cite{CaminadaS17}, and those with Abstract Dialectical Frameworks have been investigated in \cite{AlcantaraSG19}.}
Efficient mappings from AF to \emph{Answer Set Programming} (i.e. LP with \emph{Stable Model} semantics \cite{GelfondL88}) have been investigated as well \cite{SakamaR17,GagglMRWW15}.
The well-know AF system ASPARTIX 
is implemented by rewriting the input AF into an ASP program and using an ASP solver to compute extensions.
Although the ASPARTIX system allows also to reason on some extensions of AF, such as \emph{Extended AF (EAF)}~\cite{Modgil-AI-09} and AFRA, so far the relationships between LP and frameworks extending AF has not been adequately studied.
Thus, in this paper, we investigate these relationships by generalizing the work in~\cite{CaminadaSAD15}
and providing relationships between LP and different recently proposed generalizations of the Dung's framework. 
As discussed in Section~\ref{sec:conclusions},
our work is complementary to approaches
providing the semantics for an AF-based framework by flattening it into a Dung's framework.

\vspace*{2mm}
\noindent
\textbf{Contributions.}
The main contributions are as follows:
\begin{itemize}[labelsep=1mm,leftmargin=0.1in]
\setlength\itemsep{-0.0em}
\item 
We introduce a general approach for characterizing the extensions of different  AF-based frameworks under several well-known semantics in terms of Partial Stable Models (PSMs) of logic programs.
This is achieved by providing a modular definition of the sets of \textit{defeated} and \textit{acceptable} elements (i.e., arguments, attacks and supports) for each AF-based framework, 
and by leveraging on the connection between argumentation semantics and subsets of PSMs.
\item
Our approach is used to define new semantics for AFs with recursive attacks and supports under deductive  interpretation of supports, where the status of an attack is considered independently from the status of its source.
\end{itemize}

Our results can be used $i)$ for better understanding  the semantics of several AF-based frameworks, $ii)$ to easily define new semantics for extended frameworks, and $iii)$ to provide additional tools for computing stable semantics using answer set solvers \cite{ASP-Systems} and even other complete-based semantics using classical program rewriting \cite{unfolding} (see also \cite{SakamaR17,GagglMRWW15}).

\vspace*{2mm}
\noindent
\textbf{AF-based frameworks.}
It is important to observe that different frameworks extending AF share the same structure, although they have different semantics.
Thus, in the following we distinguish between framework and \textit{class} of frameworks.
Two frameworks sharing the same syntax (i.e. the structure) belong to the same (syntactic) class. For instance, BAF is a syntactic class, whereas 
AFN 
and AFD 
are two specific frameworks sharing the same BAF syntax; their semantics differ because they interpret supports in different ways.
Regarding the class \emph{Recursive AF (Rec-AF)}, where AFs are extended by allowing recursive attacks, two different frameworks called AFRA and RAF, differing only in the  determination of the status of attacks, have been proposed.
The frameworks ASAF and RAFN are two different frameworks belonging to the same class, called \emph{Recursive BAF} (Rec-BAF), consisting in the extension of BAF with recursive attacks and supports.
The differences between ASAF and RAFN semantics are not in the way they interpret supports (both based on the necessity interpretation), but in a different determination of the status of attacks as they extend AFRA and RAF, respectively.

Figure \ref{fig:overview} overviews the frameworks extending AF studied in this paper. 
Horizontal arrows denote the addition of supports with two different semantics (necessary semantics in the left direction and deductive semantics in the right direction), whereas vertical arrows denote the extension with recursive interactions (i.e., attacks and supports); the two directions denote two different semantics proposed in the literature {for determining the acceptance status of attacks}. 
Frameworks AFRAD and RAFD (in red) are novel and generalize some previously proposed frameworks.
More specifically, as shown in the figure, AFRAD (resp., RAFD) generalizes AFRA and AFD (resp., RAF and AFD), as the latters are special cases of the formers, respectively.
Clearly, frameworks in the corners are the most general ones. However, for the sake of presentation, before considering the most general frameworks, we also analyze the case of BAFs.

\begin{figure}\label{fig:overview}
\hspace*{-4mm}
\raisebox{3mm}{
    \begin{minipage}[b]{0.35\linewidth}
        \centering
        \includegraphics[width=1.25\textwidth]{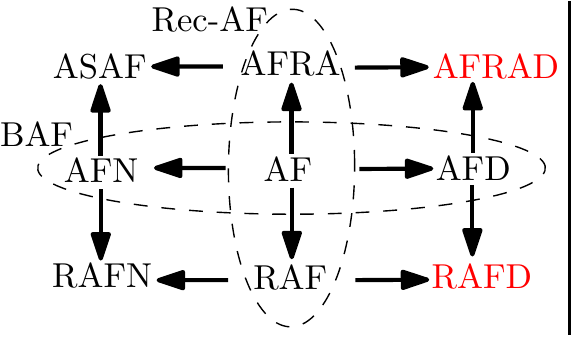}
    \end{minipage}}
    \hspace*{+1mm}
    \raisebox{3mm}{
    \begin{minipage}[b]{0.64\linewidth}
    \footnotesize
        \centering
        Legend of Acronyms:
        \vspace*{-1mm}
        \begin{itemize}
        \itemsep=0pt
        \item[] ~{\bf AF:} abstract Argumentation Framework [Dung,1995]
        \item[] ~{\bf BAF:} Bipolar AF
        		\item[] ~~~{\bf AFN:} AF with Necessities [Nouia and Risch,2011]
        		\item[] ~~~{\bf AFD:} AF with {D}eductive supports [Villata et al.,2012]
        	\item[] ~{\bf Rec-AF:} Recursive-AF
        		\item[] ~~~{\bf AFRA:} AF with Recursive Attacks [Baroni et al.,2011]
        		\item[] ~~~{\bf RAF:} Recursive AF [Cayrol et al.,2017]
        	\item[] ~{\bf Rec-BAF:} Recursive-BAF 
        		\item[] ~~~{\bf ASAF:} Attack-Support AF [Gottifredi et al.,2018]
        		\item[] ~~~{\bf RAFN:} Recursive AF with Necessities [Cayrol et al.,2018]
        		\item[] ~~~{\bf AFRAD:} AF with Rec. Attacks and Deductive supports
        		\item[] ~~~{\bf RAFD:} Recursive AF with Deductive supports
        	      	
        \end{itemize}
    \end{minipage}}
    \hspace*{-5mm}
    \centering
    \caption{AF-based frameworks investigated in the paper.}\label{fig:overview}
\end{figure}

\section{Preliminaries}\label{sec:preliminaries}
We start by recalling abstract argumentation frameworks
in increasing order of the number of features they can model. Hereafter, we will use $\mathfrak{F}$ to denote the set of the $9$ frameworks shown on left-hand side of Figure~\ref{fig:overview}. Moreover, with a little abuse of notation, we will use the same symbol $\Delta$ to denote any framework in $\mathfrak{F}$.

\subsection{Argumentation Frameworks}\label{sec:AF}

An abstract \textit{Argumentation Framework} (AF) is a pair \AFNew, 
where $\AFargsNew$ is a set of \textit{arguments} and $\AFattRelNew \subseteq \AFargsNew \times \AFargsNew$ is a set of \textit{attacks}.
An AF can be seen as a directed graph, whose nodes represent arguments and edges represent attacks;  an attack $(a,b) \in \Omega$ from $a$ to $b$ is represented by $a \rightarrow b$.

Different semantics notions have been defined leading to the characterization of collectively acceptable sets of arguments, called \emph{extensions}~\cite{Dung95}.
Given an AF $\Delta=$\AFNew  and a set $\Set{S} \subseteq \AFargsNew$ of arguments, 
an argument $a \in \AFargsNew$ is said to be
\textit{i})	
\emph{defeated} w.r.t. $\Set{S}$ iff $\exists b \in \Set{S}$  such that 
$(b, a) \in  \Omega$, and 
\textit{ii})
\emph{acceptable} w.r.t. $\Set{S}$ iff for every argument 
$b \in  \AFargsNew$ with $(b, a) \in \Omega$, there is
$c \in \Set{S} $ such that $(c,b) \in \Omega$.
The sets of defeated and acceptable arguments w.r.t. $\Set{S}$ are
defined as follows (where $\Delta$ is understood):

\vspace*{1mm}
\noindent
$\bullet$ $Def(\Set{S}) = \{ a \in A\ |\ \exists\ b \in \Set{S} \ldot (b,a) \in \Omega \}$;\\
\noindent
$\bullet$ $Acc(\Set{S})\ = \{ a \in A\ |\ \forall\ b \in A \ldot (b,a) \in \Omega\ \imply\ b \in Def(\Set{S}) \}$.

\vspace*{1mm}
\noindent 
Given an AF \AFNew, a set $\Set{S} \subseteq \AFargsNew$ of arguments is said to be $i)$
\emph{conflict-free} \iff $\Set{S}\cap Def(\Set{S}) = \emptyset$, and $ii)$ \emph{admissible} \iff it is conflict-free and $\Set{S} \subseteq Acc(\Set{S})$.

Given an AF \AFNew, 
a set $\Set{S} \subseteq \AFargsNew$ 
 is an \emph{extension} called:
\begin{itemize}[labelsep=1mm,leftmargin=0.1in]
\setlength\itemsep{-0.1em}
\item
\emph{complete} \iff it is conflict-free and $\Set{S} = Acc(\Set{S})$;
\item
\emph{preferred} iff it is a maximal $($\wrt $\subseteq)$ complete extension;
\item 
\emph{stable} \iff it is a total preferred extension, i.e. a preferred extension s.t. $\Set{S} \cup Def(\Set{S}) = A$;
\item
\emph{semi-stable} \iff it is a preferred extension 
such that $\Set{S} \cup Def(\Set{S})$ is maximal;
\item
\emph{grounded} \iff it is the smallest $($\wrt $\subseteq)$ complete extension;
\item
\emph{ideal} \iff it is the biggest $($\wrt $\subseteq)$ complete extension contained in every preferred extension.
\end{itemize}
The set of complete (resp., preferred, stable, semi-stable, grounded, ideal) extensions of a framework $\Delta$ will be denoted by ${\cal CO}(\Delta)$ (resp., ${\cal PR}(\Delta)$, ${\cal ST}(\Delta)$, ${\cal SST}(\Delta)$, ${\cal GR}(\Delta)$, ${\cal ID}(\Delta)$).

\begin{example}\label{ex:prel-af}\rm
Let $\Delta=\< A,\Omega\>$ be an AF where $A=\tt \{ a, b, c, d \}$ and $\Omega=\{\tt (a,b), (b,a), (a,c), (b,c), (c,d),$ $\tt (d,c) \}$. The set of complete extension is ${\cal CO}(\Delta)=\tt  \{ \emptyset, \{ d \}, \{ a, d \}, \{ b, d \} \}$. 
Consequently,  
${\cal PR}(\Delta)={\cal ST}(\Delta)={\cal SST}(\Delta)=\tt  \{ \{ a, d \}, \{ b, d \} \}$, ${\cal GR}(\Delta)=\tt  \{ \emptyset \}$, ${\cal ID}(\Delta)= \tt  \{ \{ d \} \}$.~\hfill~$\square$
\end{example}

\subsection{Bipolar Argumentation Frameworks}
A \textit{Bipolar Argumentation Framework} (BAF) is a triple $\< A, \Omega, \Gamma\>$, where $A$ is a set of \textit{arguments}, $\Omega \subseteq A \times A$ is a set of \textit{attacks}, and $\Gamma \subseteq A \times A$ is a set of \textit{supports}.
A BAF can be  represented by a directed graph with two types of edges: \emph{attacks} and \emph{supports}, denoted by $\rightarrow$ and $\Rightarrow$, respectively. 
A \emph{support path} $a_0 \supportplus a_n$ from argument $a_0$ to argument $a_n$ is a  sequence of $n$ edges $a_{i-1} \Rightarrow a_i$ with $0 < i \leq n$.
We use $\Gamma^+ =$ $\{ (a,b) \ |$ $a,b \in A \ \wedge a \supportplus b \}$ to denote the set of pairs $(a,b)$ such that there exists a support path from $a$ to $b$.
It is assumed that $\Gamma$ is acyclic.

Different interpretations of the support relation have been proposed~\cite{Rahwan-Simari-Book,CayrolL13,CohenGGS14}. 
Given a BAF $\Delta$ 
and an interpretation $\cal I$ of the support relation, the semantics of $\Delta$ w.r.t. $\cal I$ can be given in terms of an equivalent AF $\Delta_{\cal I}$, 
derived from $\Delta$ by substituting supports 
with the so-called \emph{complex} or \emph{extended} attacks.
In this paper we consider $\I\in\{d,n\}$, where $d$ and $n$ denote \emph{deductive} and \emph{necessary} interpretation of supports proposed in \cite{VillataBGT12} and  \cite{NouiouaR11}, respectively.

\vspace*{1mm}
\noindent
{\bf AF with Necessities (AFN).}
An AFN is a BAF where supports are interpreted as necessary.
The necessary interpretation of a support $a \Rightarrow b$ is that $b$ is accepted only if $a$ is accepted. 
Given an AFN $\Delta = \<A,\Omega,\Gamma\>$, there exists an \emph{extended} \emph{attack} from $a$ to $b$ if there are either: \\
$\bullet$
an attack $a \rightarrow c$ and a support path $c \supportplus b$ (that we call \emph{supported attack}), or\\
$\bullet$ 
a support path $c \supportplus a$ and an attack $c \rightarrow b$ (that we call \emph{mediated attack}). \\
We denote by $\Delta_n = \<A,\Omega_n\>$ the AF derived from $\Delta$ by replacing supports with extended attacks.

\vspace*{1mm}
\noindent
{\bf AF with Deductive supports (AFD).}
An AFD is a BAF where supports are interpreted as deductive.
The deductive interpretation of a support $a \Rightarrow b$  is that $b$ is accepted whenever $a$ is accepted (and $a$ is defeated whenever $b$ is defeated). 
Given an AFD $\Delta = \<A,\Omega,\Gamma\>$, there exists a \emph{complex attack} from argument $a$ to argument $b$ if there are either: \\
$\bullet$ a support path $a \supportplus c$ and an attack $c \rightarrow b$ (\emph{supported attack}), or \\
$\bullet$ 
an attack $a \rightarrow c$ and support path $b \supportplus c$ (\emph{mediated attack}).

\noindent
$\Delta_d = \<A,\Omega_d\>$ denotes the AF derived from $\Delta$ by replacing supports with complex attacks.

Given a BAF $\<A, \Omega, \Gamma\>$ with interpretation ${\cal I} \in \{n,d\}$ of supports, and a set of arguments $\Set{S}\subseteq A$, then 
$Def(\Set{S}) = \{ a \in A\ \mid\ \exists b \in \Set{S} \ldot (b,a) \in \Omega_{\cal I} \}$, and  
$Acc(\Set{S}) = \{a \in A\ \mid\ \forall b \in A\ldot(b,a) \in \Omega_{\cal I}\imply~b \in Def(\Sb)\}$.

\begin{example}\label{ex:prel-baf-extended-attack}\rm
Consider the BAF $\Delta = \<A,\Omega, \Gamma\>$ of Figure~\ref{fig:asaf-one}. 
Under the necessary interpretation of supports $\Delta_n = \<A, \Omega_n\>$, where 
$\Omega_n = \tt \{ (w_i,r), (w_i,w_e), (w_e,p) \}$.
$\Delta_n$ has a unique complete extension $\tt \{ w_i,p\}$.
Dually, under the deductive interpretation of supports $\Delta_d = \<A, \Omega_d\>$, where 
$\Omega_d = \tt \{ (w_i,r), (r,p), (w_e,p) \}$.
$\Delta_d$ has a unique complete extension $\tt \{ w_i,w_e\}$.~\hfill~$\square$
\end{example}
\subsection{Recursive Argumentation Frameworks}\label{sec:AFRApreliminaries}
A \textit{Recursive Argumentation Framework (\reaf)} is a tuple $\< A, \Sigma, {\bf s}, {\bf t}\>$, where $A$ is a
set of arguments, $\Sigma$ 
is a set disjunct from $A$ representing attack names, ${\bf s}$ (resp., ${\bf t}$) is a function from $\Sigma$ to $A$ (resp., to $(A \cup \Sigma)$) mapping each attack to its source (resp., target).
An attack may be recursive as an argument may attack an argument 
or an attack, and extensions may contain both arguments and attacks.
Two different semantics have been proposed in literature.

\vspace*{1mm}
\noindent
{\bf Recursive AF (RAF).}
In \cite{CayrolFCL17} a semantic framework for Rec-AF, called \textit{Recursive Argumentation Framework}, is proposed.
The semantics for an RAF is given in terms of \emph{defeated} and \emph{acceptable} sets.

\vspace*{1mm}
\noindent
$\bullet\ {Def}(\Set{S}) = \{ X \in A \cup \Sigma\ |\ \exists\ \alpha \in \Sigma \cap \Sb \ldot {\bf s}(\alpha) \in A\cap\Sb \wedge {\bf t}(\alpha)=X\}$;

\vspace*{1mm}
\noindent
$\bullet\ Acc(\Set{S})\, = \{ X \in A \cup \Sigma\ |\ 
\forall \alpha \in \Sigma \ldot {\bf t}(\alpha) = X \imply 
 \alpha \in Def(\Sb) \vee {\bf s}(\alpha) \in Def(\Sb) \}$.

\vspace*{1mm}
{The peculiarity of RAF semantics is that an attack is defeated only if it is explicitly attacked and, consequently, can be accepted whenever its source is defeated.}

\vspace*{2mm}
\noindent
{\bf AF with Recursive Attacks (AFRA).}
{Differently from RAF semantics, in an AFRA \cite{Baroni-Cerutti-Giacomin-Guida11} the status of an attack is also related to the status of its source argument.}

Given $X \in A \cup \Sigma$ and $\alpha \in \Sigma$, we say that $\alpha$  \emph{(directly or indirectly) attacks} $X$ (denoted by $\alpha\ {\tt def}\ X$) if either ${\bf t}(\alpha) = X$ or ${\bf t}(\alpha) = {\bf s}(X)$.
Given an AFRA $\<A, \Sigma, {\bf s},{\bf t}\>$\footnote{ For the sake of presentation, we consider a slight generalization of AFRA, where attack names are first-class citizens, allowing to also represent more than one attack from the same source to the same target. 
In the original work an AFRA is a tuple $\<A,\Omega\>$ where A is a set of arguments and $\Omega$ is a set of attacks $\Omega: A\rightarrow (A\cup \Omega)$~\cite{Baroni-Cerutti-Giacomin-Guida11}.} and a set  $\Set{S}\subseteq A \cup \Sigma$
of arguments and attacks, 
the \emph{defeated} and \emph{acceptable} sets are:

\vspace*{1mm}
\noindent
$\bullet$
$Def(\Set{S}) = \{ X \in A \cup \Sigma\ |\ \exists\ \alpha \in \Sigma \cap \Sb \ldot \alpha\ {\tt def}\ X \}$;

\noindent
$\bullet\ Acc(\Set{S}\,) = \{ X \in A\ \cup\ \Sigma\ | \
	\forall\ \alpha \in \Sigma \ldot \alpha\ {\tt def}\ X \imply \!\alpha \in Def(\Sb)  \}$.

\vspace*{1mm}
\noindent 
The idea behind AFRA semantics is that whenever an argument $a$ is defeated, every attack starting from $a$ is (indirectly) defeated as well. 

The notions of \emph{conflict-free}, \emph{admissible sets}, and 
the different types of extensions can be defined in a standard way (see Section \ref{sec:AF}) by considering $\Set{S}\subseteq A \cup \Sigma$
and by using the new  definitions of defeated and acceptable sets reported above.

\begin{example}\label{ex:prel-afra-raf}\rm
Let $\Delta=\< A,\Sigma, {\bf s},{\bf t}\>$ be an Rec-AF, where $A=\{\tt a,b,c\}$, $\Sigma=\{\tt \alpha_1, \alpha_2 \}$, ${\bf s} = \{\tt \alpha_1/a, \alpha_2/b \}$, ${\bf t} = \{\tt \alpha_1/b, \alpha_2/c \} \>$ where $\alpha/y \in {\bf  s}$ (resp., $\beta/y \in {\bf  t}$) denotes that ${\bf s}(\alpha)=y$ (resp., ${\bf t}(\beta)=y$).
Considering the set $\Sb =\tt \{\tt a, \alpha_1 \}$, under the AFRA (resp., RAF) semantics we have that $Def(\Sb) =\tt \{\tt b, \alpha_2 \}$ (resp., $Def(\Sb) =\tt \{\tt b \}$), and there exists a unique complete extension $\tt \{ a, c, \alpha_1 \}$ (resp., $\tt \{ a, c, \alpha_1, \alpha_2 \}$).~\hfill~$\square$
\end{example}

It has been shown that RAF and AFRA semantics may differ only in the status of attacks, and extensions under RAF semantics could be derived from extensions under AFRA semantics and vice versa~\cite{CayrolFCL17}.

\subsection{Recursive Bipolar Argumentation Frameworks with Necessities}\label{sec:ASAFpreliminary}
By combining the concepts of both bipolarity and recursive interactions, more general argumentation frameworks have been defined.

A \textit{Recursive Bipolar Argumentation Framework (Rec-BAF)} is a tuple $\< A, \Sigma, \Pi, {\bf s}, {\bf t} \>$, where $A$ is a set of arguments, $\Sigma$ is a set of attack names, $\Pi$ is a set of necessary support names, ${\bf s}$ (resp., ${\bf t}$) is a function from $\Sigma\ \cup\ \Pi$ to $A$ (resp., to $A\ \cup\ \Sigma\ \cup\ \Pi$) mapping each attack/support to its source (resp., target). 
In the following, given a set $\Phi$ such that either $\Phi \subseteq \Sigma$ or $\Phi \subseteq \Pi$, we denote by $i)$ $\Phi^* = \{ (\sb(\gamma),\tb(\gamma)) \mid \gamma \in \Phi \}$ the set of pairs connected by an attack/support edge, and $ii)$ $\Phi^+$ the transitive closure of $\Phi$. It is assumed that $\Pi^*$ is acyclic. 

Two different semantics have been defined
 under necessary interpretation of supports.

\vspace*{1mm}
\noindent
{\bf Recursive AF with Necessities (RAFN).}
The \emph{Recursive Argumentation Framework with Necessities} has been proposed in~\cite{CayrolFCL18}. The semantics combines the RAF interpretation of attacks with that of BAF under the necessity interpretation of supports~(i.e., AFN). Here we consider a simplified version where supports have a single source and the support relation is acyclic.
Formally, given an RAFN $\< A, \Sigma, \Pi, {\bf s}, {\bf t} \>$, $X \in (A \cup \Sigma \cup \Pi)$, $a \in A$, and $\Sb\subseteq A \cup \Sigma \cup \Pi$, we say that argument $a$
\emph{recursively attacks $X$ given $\Sb$} (denoted as $a\ {\tt att_{\Sb}}\ X$)  if either $(a,X)\in (\Sigma\cap \Sb)^*$ or there exists $b \in A$ such that {$(a,b)\in (\Sigma\cap \Sb)^*$} and $(b,X) \in (\Pi \cap \Sb)^+$. 

For any RAFN $\Delta$ and $\Set{S}\subseteq A \cup \Sigma \cup \Pi$, the \emph{defeated} and \emph{acceptable} sets (given $\Sb$) are:

\vspace*{1mm}
\noindent
$\bullet\ Def(\Set{S}) = \{ X \in A \cup \Sigma \cup \Pi\ |\ 
	\exists b \in A \cap \Sb \ldot 
	b\ {\tt att_\Sb}\ X\}$; 

\noindent
$\bullet\ Acc(\Set{S})\, = \{ X \in A \cup \Sigma \cup \Pi\ |\ 
	\forall b \in A \ldot b\ {\tt att_\Sb}\ X \imply b \in  Def(\Sb)\}$.

\vspace*{2mm}
\noindent
{\bf Attack-Support AF (ASAF).}
The \emph{Attack-Support Argumentation Framework (ASAF)} has been proposed in \cite{CohenGGS15,Gottifredi-Cohen-Garcia-Simari18}.
The semantics combines the AFRA interpretation of attacks with that of BAF under the necessary interpretation of supports (i.e., AFN). For the sake of presentation, we consider a slight generalization of ASAF, where attack and support names are first-class citizens, giving the possibility to represent multiple attacks and supports from the same source to the same target.\footnote{In the original work \cite{CohenGGS15,Gottifredi-Cohen-Garcia-Simari18} an ASAF is a tuple $\<A,\Omega,\Gamma\>$ where A is a set of arguments, $\Omega$ is a set of attacks
$\Omega: A\rightarrow (A\cup \Omega)$, and $\Gamma$ is a set of supports 
$\Gamma: A\rightarrow (A\cup \Gamma)$.}

Formally, given an ASAF $\< A, \Sigma, \Pi, {\bf s}, {\bf t} \>$, $X \in (A \cup \Sigma \cup \Pi)$, $\alpha \in \Sigma$, and $\Sb\subseteq A \cup \Sigma \cup \Pi$, we say that $i)$ 
$\alpha$  \emph{(directly or indirectly) attacks} $X$ (denoted by $\alpha\ {\tt def}\ X$) if either ${\bf t}(\alpha) = X$ or ${\bf t}(\alpha) = {\bf s}(X)$, and $ii)$ $\alpha$
\emph{extendedly defeats $X$ given $\Sb$} (denoted as $\alpha\ {\tt def_{\Sb}}\ X$)  if either $\alpha\ {\tt def}\ X$ or there exists  $b \in A$ such that ${\bf t}(\alpha) = b$ and either  $(b,X) \in (\Pi \cap \Sb)^+$ or $(b,{\bf s}(X)) \in (\Pi \cap \Sb)^+$. 
For any ASAF $\Delta$ and $\Set{S}\subseteq A \cup \Sigma \cup \Pi$, the \emph{defeated} and \emph{acceptable} sets (given $\Sb$) are:

\vspace*{1mm}
\noindent
$\bullet\ Def(\Set{S}) = \{ X \in A \cup \Sigma \cup \Pi\ |\ 
	\exists\ \alpha \in \Sigma \cap \Sb \ldot 
	\alpha\ {\tt def_\Sb}\ X\}$; 

\noindent
$\bullet\ Acc(\Set{S})\, = \{ X \in A \cup \Sigma \cup \Pi\ |\ 
	\forall \alpha \in \Sigma \ldot \alpha\ {\tt def_\Sb}\ X \imply \alpha \in Def(\Sb) \}$.

\vspace*{1mm}
Again, the notions of \emph{conflict-free}, \emph{admissible sets}, and the different types of extensions can be defined in a standard way (see Section \ref{sec:AF}) by considering $\Set{S}\subseteq A \cup \Sigma \cup \Pi$
and by using the definitions of defeated and acceptable sets reported above.

Note that for AFs with high-order interactions the mapping to AF is not trivial, as in the case of BAF, because extensions also contain attacks and supports.
\blue{In particular, an equivalent AF for an ASAF can be obtained by translating it into an AFN~\cite{CohenGGS15} that in turns 
can be translated into an  
AF~\cite{NouiouaR11} (see also~\cite{Gottifredi-Cohen-Garcia-Simari18}).
}

\begin{example}\label{ex:prel-afras}\rm
Consider the Rec-BAF $\Delta$ with necessary supports of Figure~\ref{fig:asaf-two}.
Under both ASAF and RAFN semantics ${\cal CO}(\Delta)=\{\{\tt w_i,$ $\tt r,$ $\tt w_e,w_t,$ $\tt \alpha_2, \alpha_3, \beta_1 \}\}$.
Consider now the Rec-BAF $\Delta'$ obtained by adding to $\Delta$ an argument $\tt s$ attacking argument $\tt w_t$ with attack $\alpha_4$.
Under the ASAF semantics $\Delta'$ has a unique complete extension $\{\tt w_i, s, p,$\ $\tt \alpha_1, \alpha_4, \beta_1\}$; note that attacks $\tt \alpha_2$  and $\alpha_3$ are not part of the extension as their sources (i.e., $\tt w_e$ and $\tt w_t$, respectively) are defeated.
Differently,  
$\{\tt w_i, s, p,$ $\tt \alpha_1, \alpha_2, \alpha_3,$ $\tt \alpha_4, \beta_1\}$ is the only complete extension of $\Delta'$ under the RAFN semantics. 
~\hfill~$\square$
\end{example}

Analogous to the case of Rec-AFs, ASAF and RAFN semantics may differ only in the status of attacks. Moreover, for each semantics, the RAFN extensions can be derived from the corresponding ASAF extensions and vice versa.

\subsection{Partial Stable Models}
We summarize the basic concepts which underly the notion of PSMs \cite{SaccaZ90}. 
  
A (normal, logic) program is a set of rules of the form
$A \leftarrow B_1 \wedge \cdots \wedge B_n$, with $n \geq 0$, where $A$ is an atom, called head, and $B_1\wedge \cdots \wedge B_n$ is a conjunction of literals, called body. 
We consider programs without function symbols.
Given a program $P$, $ground(P)$ denotes the set of all ground instances of the rules in $P$. 
The Herbrand Base of a program $P$, i.e. the set of all ground
atoms which can be constructed using predicate and constant symbols occurring in $P$, is denoted by $B_P$, whereas $\neg B_P$ denotes the set $\{ \neg A \mid A \in B_P \}$. 
Analogously, for any set $S \subseteq B_P \cup \neg B_P$, $\neg S$ denotes the set $\{ \neg A \mid A \in S \}$, where $\neg \neg A = A$.
Given $I \subseteq B_P \cup \neg B_P$, $pos(I)$ (resp., $neg(I)$) stands for $I \cap B_P$ (resp., $\neg I \cap B_P$).
$I$ is \emph{consistent} if $pos(I) \cap \neg neg(I) = \emptyset$, otherwise $I$ is \emph{inconsistent}.

Given a program $P$, $I \subseteq B_P \cup \neg B_P$
is an \emph{interpretation}  of $P$ if $I$ is consistent. 
Also, $I$ is \emph{total} if $pos(I) \cup  neg(I) = B_P$, \emph{partial} otherwise.
A partial interpretation $M$ of a program $P$ is a
\emph{partial model} of $P$ if for each $\neg A \in M$ every rule in $ground(P)$ having as head $A$ contains at least one body literal $B$ such that $\neg B \in M$.
Given a  program $P$ and a partial model $M$, the positive instantiation of $P$ w.r.t. $M$, denoted by $P^M$, is obtained from $ground(P)$ by deleting:\
$(a)$
each rule containing a negative literal $\neg A$ such that $A \in pos(M)$;\
$(b)$\
each rule containing a literal $B$ such that neither $B$ nor $\neg B$ is in $M$;\
$(c)$
all the negative literals in the remaining rules.
Clearly, all the rules in $P$ are definite clauses and hence the minimal Herbrand model of $P$ can be obtained as the least fixpoint of its immediate consequence operator $T_{P^M}$, denoted by $T_{P^M}^\omega(\emptyset)$. 
For any partial model $M$ of a logic program $P$,  $T_{P^M}^\omega(\emptyset) \subseteq M$ \cite{SaccaZ90}.

Let $P$ be a program and $M$ a partial model for $P$. 
Then $M$ is  \
$(a)$\	
\emph{founded} if $T_{P^M}^\omega(\emptyset) = pos(M)$; \
$(b)$\	
\emph{stable} if it is founded and it is not a proper subset of any other founded
model.
The set of partial stable models of a logic program $P$, denoted by ${\cal PM}(P)$, define a meet semi-lattice. The \textit{well-founded} model (denoted by ${\cal WF}(P)$) and the \textit{maximal-stable} models ${\cal MS}(P)$\footnote{Corresponding to the \emph{preferred extensions} of \cite{Dung91}.}, 
are defined by considering $\subseteq$-minimal and $\subseteq$-maximal elements. 
The set of (total) \textit{stable} models (denoted by ${\cal SM}(P)$) is obtained by considering the maximal-stable models which are total, 
whereas the \textit{least-undefined} models (denoted by ${\cal LM}(P)$) are obtained by considering the maximal-stable models with a $\subseteq$-minimal set of undefined atoms (i.e., atoms which are neither true or false).
The \textit{max-deterministic} model (denoted by ${\cal MD}(P)$) is the $\subseteq$-maximal PSM contained in every maximal-stable model \cite{Sacca97,GrecoS99}.

\begin{example}\label{ex:prel-af}\rm
Consider the program $P$ consisting of the following four rules $\{\tt a \leftarrow \neg b;\ b \leftarrow \neg a;\ c \leftarrow \neg a \wedge \neg b \wedge \neg d;\ d \leftarrow \neg c \}$.
The set of partial stable models  of $P$  is ${\cal PS}(P)=\tt  \{\ \emptyset,\ \{ \neg c, d \},$ $\tt \{ a, \neg b, \neg c, d \},\ \{ \neg a, b, \neg c, d \}\ \}$. 
Consequently, 
${\cal WF}(\Delta)=\tt  \{\ \emptyset\ \}$, 
${\cal MD}(\Delta)= \tt  \{\ \{ \neg c, d \}\ \}$, 
${\cal MS}={\cal ST}(\Delta)={\cal LS}(\Delta)=\tt  \{ \{ a, \neg b, \neg c, d \}, \{ \neg a, b, \neg c, d \} \}$.~\hfill~$\square$
\end{example}

\paragraph{Propositional Programs.}

Given a set of symbols $\Lambda = \{a_1, ..., a_n\}$, 
a \textit{(propositional) program} over $\Lambda$ is a set of $|\Lambda|$ rules $a_i \leftarrow body_i$ ($1\leq i\leq n$), where every $body_i$ is a propositional formula defined over $\Lambda$.   	
The semantics of a propositional program $P$, defined over a given alphabet $\Lambda$, is given in terms of the set ${\cal PS}(P)$ of its Partial Stable Models (PSMs) that are obtained as follows: \textit{i}) $P$ is first rewritten into a set of standard (ground) logic rules $P'$, whose bodies contain conjunction of literals (even by adding fresh symbols to the alphabet)\footnote{A rule $a \leftarrow (b \vee c) \wedge (d \vee e)$ is rewritten as $a \leftarrow \neg a_1  \wedge \neg a_2$, \ \ $a_1 \leftarrow \neg b \wedge \neg c$ \ and \ $a_2 \leftarrow \neg d \wedge \neg e$.}; 
\textit{ii}) next, the set of PSMs of $P'$ is computed;
\textit{iii}) finally, fresh literals added to $\Lambda$ in the first step are deleted from the models. 
It is worth noting that for propositional programs we can assume as Herbrand Base the set of (ground) atoms occurring in the program.

\section{A Logic Programming Approach}
In this section we present a new way to define the semantics of AF-based frameworks by considering propositional programs and partial stable models.
In order to compare extensions $E$ of a given framework $\Delta$ (containing acceptable elements) with PSMs of a given program $P$ (containing true and false atoms), we denote as $\widehat{E} = E \cup \{ \neg a \mid a \in Def(E) \}$ the \emph{completion} of $E$. Moreover, for a collection of extensions $\Set{E}$, $\widehat{\Set{E}}$ denotes the set $\{ \widehat{E} \mid E \in \Set{E} \}$.

Observe also that for any framework $\Delta$ and complete extension $E$ for $\Delta$, elements not occurring in $E \cup Def(E)$ are said to be undecided (or undefined), whereas for any program $P$ and PSM $M$ for $P$, atoms not occurring in $pos(M) \cup neg(M)$ are said to be undefined.
Thus, to compare complete extensions and PSMs it is sufficient to consider the completion of extensions.

The next proposition states the relationship between the argumentation frameworks (e.g. AF, BAF, Rec-AF, etc.) and logic programs with partial stable models.

\begin{proposition}\label{pro:proposition1}
For any framework $\Delta\in \mathfrak{F}$ and a propositional program $P$,
whenever $\widehat{{\cal CO}(\Delta)} = {\cal PS}(P)$  it holds that $\widehat{{\cal PR}(\Delta)} = {\cal MS}(P)$, $\widehat{{\cal ST}(\Delta)} = {\cal ST}(P)$, $\widehat{{\cal SST}(\Delta)} = {\cal LM}(P)$, $\widehat{{\cal GR}(\Delta)} = {\cal WF}(P)$, and $\widehat{{\cal ID}(\Delta)} = {\cal MD}(P)$.\footnote{For the novel frameworks $\Delta\in\{$AFRAD, RAFD$\}$, the set ${\cal CO}(\Delta)$ of the complete extensions,
and the sets of extensions prescribed by the other semantics,
are defined in 
Section~\ref{sec:NovelSemanticsAFRS}.}
\end{proposition}

The result of Proposition~\ref{pro:proposition1} derives from the fact that preferred, stable, semi-stable, grounded, and ideal extensions are defined by selecting a subset of the complete extensions satisfying given criteria (see Section \ref{sec:preliminaries}). 
On the other~side,~the maximal, stable, least-undefined, well-founded, and max-deterministic (partial) stable models are obtained by selecting a subset of the PSMs satisfying criteria coinciding with those used to 
restrict the set of complete extensions.

Given a framework $\Delta$ and an extension $E$, for any element $a$ which could occur in some extension of $\Delta$, the truth value $v_{_E}(a)$, or simply $v(a)$ whenever $E$ is understood, is equal to $\true$ if $a \in E$, $\false$ if $a \in Def(E)$, $\undec$ (\emph{undecided}) otherwise.
Hereafter, we assume that $\false < \undec < \true$ and $\neg \undec = \undec$.

The strict relationship between the semantics of AFs (given in terms of subset of complete extensions) and the semantics of logic programs (given in terms of subset of PSMs) has been shown 
\blue{in~\cite{WuCG09,CaminadaSAD15}.} 
The relationship is based on the observation that the meaning of an attack $a \rightarrow b$ is that the condition $v(b) \leq \neg v(a)$ must hold.
On the other side, the satisfaction of a logical rule $a \leftarrow b_1,...,b_n$ implies that $v(a) \geq {min}\{v(b_1),...,v(b_n),\true\}$. 

\begin{definition}\label{def:af-prog}
Given an AF $\Delta = \<A, \Omega\>$, we denote as
$P_\Delta = \{ a \leftarrow \bigwedge_{(b,a) \in \Omega} \neg b\ |\ a \in A \}$ the propositional 
program derived from $\Delta$.
\end{definition}

The semantics of an AF $\Delta$ can be obtained by considering PSMs of the logic program $P_\Delta$. Particularly, for any AF $\Delta$,  $\widehat{{\cal CO}(\Delta)} = {\cal PS}(P_{\Delta} )$.
Therefore, a natural question is:
\emph{Can we also model semantics defined for frameworks extending AF by means of PSMs of logic programs?} The answer is \emph{Yes} and we shall investigate this relationship in the rest of the paper. 

Although for a BAF $\Delta$ with deductive (resp., necessary) supports this could be carried out by considering the program $P_{\Deltad}$ (resp., $P_{\Deltan}$), where $\Deltad$ (resp., $\Deltan$) is the AF obtained from $\Delta$ by substituting supports with complex (resp., extended) attacks, we propose a general method that can be applied to all the discussed frameworks, and even to new frameworks (see Section~\ref{sec:NovelSemanticsAFRS}).

In order to model frameworks extending Dung's framework by logic programs under PSM semantics, we provide new definitions of defeated and acceptable sets that, for a given set $\Set{S}$, will be denoted by $\Def(\Set{S})$ and $\Acc(\Set{S})$, respectively.
These definitions will be used to derive rules in $P_\Delta$. 
For AFs we have that for every set $\Set{S} \subseteq A$,  $\Def(\Set{S}) = Def(\Set{S})$ and $\Acc(\Set{S}) = Acc(\Set{S})$.

\subsection{Bipolar AFs}\label{subsec:bafs-programs}
To extend the above result to more general frameworks containing supports (i.e. BAFs and recursive BAFs), we need to separately consider different interpretations of supports.

\vspace*{1mm}
\noindent
{\bf AFN.}
The necessary interpretation of supports means that whenever there is a support $a \Rightarrow b$, the condition $v(b) \leq v(a)$ must hold. 
Thus, defeated and acceptable sets can be defined as follows.

\begin{definition}\label{ACC:def}
For any AFN $\<A, \Omega, \Gamma\>$ and 
set of arguments $\Set{S} \subseteq A$,

\vspace*{1mm}
\noindent
$\bullet	 \Def(\Set{S}) = \{ a \in A\ |\ (\exists b \in \Set{S} \ldot (b,a) \in \Omega)\ \vee\  (\exists c \in \Def(\Set{S}) \ldot  (c,a) \in \Gamma)  \}$;

\noindent
$\bullet	 \Acc(\Set{S})\! = \!\{ a\! \in\! A\, |\, (\forall b\! \in\! A \ldot  (b,a)\! \in \Omega \imply b\! \in\! \Def(\Set{S})) \wedge (\forall c\in A \ldot  (c,a)\! \in\! \Gamma \imply c\! \in\! \Acc(\Set{S})) \}$.
\end{definition}

It is worth noting that $\Def(\Set{S})$ and $\Acc(\Set{S})$ are defined recursively, and that in general they may differ from $Def(\Set{S})$ and $Acc(\Set{S})$, respectively, as shown in the following example.

\begin{example}\label{ex:BAF2}\rm
Let $\tt \< \{\a,\ \b,\ \c,\ \d\},\tt \{\ (\b,\c),\ (\c,\d)\ \}, \{(\b, \a) \} \>$ be\  an\  AFN. Then,  $Def(\{\a\}) = \{ \c \}$ and $Acc(\{\a\}) =$ $ \{ \a, \b, \d \}$, whereas $\Def(\{\a\}) = \emptyset$ and $\Acc(\{\a\}) =$ $ \{ \a,\b \}$. 
On the other hand $Def(\{\a, \b, \d\}) = $ $\Def(\{\a, \b, \d\}) = \{ \c \}$ and $Acc(\{\a, \b, \d\}) = \Acc(\{\a, \b, \d\}) = \{ \a, \b, \d \}$.~\hfill~$\square$
\end{example}


\begin{theorem}\label{BAFn:Acceptable-Sets-Equivalence}
Given an AFN $\Delta$ and an extension $\Set{S} \in\! {\cal CO}(\Delta)$, then $Def(\Sb) \= \Def(\Sb)$ and $Acc(\Sb) \= \Acc(\Sb)$. 
\end{theorem}

Theorem~\ref{BAFn:Acceptable-Sets-Equivalence}  states that in order to define the semantics for an AFN $\Delta$ we can use acceptable sets $\Sb = \Acc(\Sb)$. This is captured by the following definition, that shows how to derive a propositional program from an AFN.

\begin{definition}\label{def:BAFn-program}
Given an AFN $\Delta = \<A, \Omega, \Gamma\>$, then $P_\Delta = \{ a \leftarrow (\bigwedge_{(b,a) \in \Omega} \neg b\ \wedge\  
							\bigwedge_{(c,a) \in \Gamma} c)\ |$ $ a \in A \}$ denotes the propositional program derived from $\Delta$.
\end{definition}

\begin{theorem}\label{BAFn:semantics-equivalence}
For any AFN $\Delta$, \ \ $\widehat{{\cal CO}(\Delta)} = {\cal PS}(P_{\Delta} ) = {\cal PS}(P_{\Deltan})$.
\end{theorem}

The previous theorem states that the set of complete extensions of an AFN $\Delta$ coincides with the set of PSMs of the derived logic program $P_{\Delta}$. Consequently the set of PSMs of $P_{\Delta}$ and $P_{\Delta_n}$, derived from the AF $\Delta_n$, also coincide. 
Moreover, using Proposition~\ref{pro:proposition1}, also the others argumentation semantics turns out to be characterized in terms of subsets of PSMs.

\begin{example}\rm
Consider the AFN $\Delta$ of Figure~\ref{fig:asaf-one}. 
Then, the propositional program derived from $\Delta$ is $P_\Delta=\{\tt (w_i\leftarrow),\ (r\leftarrow\neg w_i),\ (w_e \leftarrow r),\  (p \leftarrow \neg w_e)\}$, 
and $P_{\Deltan}=\{\tt (w_i \leftarrow ),\ (r \leftarrow  \neg w_i),\ (w_e \leftarrow \neg w_i),\  (p \leftarrow \neg w_e )\}$.
Clearly, $\widehat{{\cal CO}(\Delta)} = {\cal PS}(P_\Delta) = {\cal PS}(P_{\Deltan}) =  \{ \{ \tt w_i, \neg r, \neg w_e, p\} \}$.~\hfill~$\square$
\end{example}

\vspace*{1mm}
\noindent
{\bf AFD.}
The deductive interpretation of supports means that whenever there is a support $a \Rightarrow b$, the condition $v(a) \leq v(b)$ must hold. 
Thus, defeated and acceptable sets can be defined as follows.

\begin{definition}
For any AFD $\Delta = \<A, \Omega, \Gamma\>$ and  set of arguments $\Set{S} \subseteq A$,

\noindent
$\bullet\Def(\Set{S}) = \{ a \in A\ |\ (\exists\ b \in \Set{S} \ldot  (b,a) \in \Omega)\ \vee\	(\exists\ c \in \Def(\Set{S}) \ldot (a,c) \in \Gamma)  \}$;

\noindent
$\bullet\Acc(\Set{S})\! =\! \{ a \!\in\! A\ |\ (\forall\ b\! \in A \ldot  (b,a)\! \in \Omega\! \imply b \in\! \Def(\Set{S}))\wedge  	(\forall c \in\! A \ldot\! (a,c)\! \in \!\Gamma \imply c \in\! \Acc(\Set{S})) \}$.
\end{definition}

\begin{theorem}\label{BAFd:Acceptable-Sets-Equivalence}
Given an AFD $\Delta$ and an extension $\Set{S} \in {\cal CO}(\Delta)$, then $Def(\Sb) \= \Def(\Sb)$ and $Acc(\Sb) \= \Acc(\Sb)$.
\end{theorem}

We derive a program $P_\Delta$ from a given AFD $\Delta$ as follows.

\begin{definition}\label{def:BAFd-program}
Given an AFD $\Delta = \<A, \Omega, \Gamma\>$, then 
$P_\Delta = \{ a \leftarrow (\bigwedge_{(b,a) \in \Omega} \neg b\ \wedge\ \bigwedge_{(a,c) \in \Gamma} c)\ |$ $a \in A \}$ denotes the propositional program derived from $\Delta$.
\end{definition}
\noindent

Similarly to what done earlier, results stating the relationships between AFD semantics and partial stable models can be obtained.

\begin{theorem}\label{BAFd:semantics-equivalence}
For any AFD  $\Delta$, $\widehat{{\cal CO}(\Delta)}= {\cal PS}(P_{\Delta} )={\cal PS}(P_{\Deltad}) $.
\end{theorem}

\subsection{Recursive BAFs with Necessary Supports}\label{sec:AFRAS}

In this section we study the relationship between partial stable models and the semantics of Rec-BAFs.
Particularly, we first present results for RAFN semantics, and then we discuss results for the ASAF framework. 
We remand to the next section the presentation of two novel semantics for recursive bipolar AFs with deductive interpretation of supports.

\vspace*{2mm}
\noindent
{\bf RAFN.} 
We next provide the definitions of defeated and acceptable sets for an RAFN.
\begin{definition}
For any RAFN $\< A, \Sigma, \Pi, {\bf s}, {\bf t} \>$ and set $\Set{S} \subseteq A \cup \Sigma \cup \Pi$, we have that: 

\vspace*{1mm}
\noindent
$\bullet\ {\Def}(\Set{S}) = \{ X \in A \cup \Sigma \cup \Pi\ |\ (\exists \alpha \in \Sigma \cap \Set{S} \ldot  \sb(\alpha) \in \Sb \wedge \tb(\alpha)=X) \ \vee \\ 
\hspace*{+46mm}(\exists~\beta~\in~\Pi \cap~\Set{S} \ldot \sb~(\beta)~\in~\Def(\Sb)~\wedge~\tb(\beta)=X)\ \}$;

\vspace*{1mm}
\noindent
$\bullet\ \Acc(\Set{S})\! =\!\{ X \in A\cup \Sigma\cup \Pi\ |\ (\forall \alpha\! \in\! \Sigma  \ldot \tb(\alpha)=X \imply\ (\alpha \in \Def(\Set{S}) \vee {\bf s}(\alpha) \in\! \Def(\Set{S})))\wedge \\
\hspace*{+46mm}  (\forall \beta\! \in\! \Pi \ldot \tb(\beta)=X\! \imply\!    (\beta \in \Def(\Set{S}) \vee {\bf s}(\beta) \in \Acc(\Set{S})))\ \}$.
\end{definition}

The following theorem allows to easily derive the propositional program for any RAFN, by directly looking at the set $\Acc(\Sb)$ of acceptable elements.

\begin{theorem}\label{RAFN:Acceptable-Sets-Equivalence}
Given an RAFN $\Delta$ and an extension $\Set{S} \in {\cal CO}(\Delta)$, then $Def(\Sb) \= \Def(\Sb)$ and $Acc(\Sb) \= \Acc(\Sb)$.
\end{theorem}

\begin{definition}\label{def:AFRAS-program-new-sem}
Given an RAFN 
$\Delta = \<A, \Sigma, \Pi, {\bf s}, {\bf t}\>$, then 
$P_\Delta $ (the propositional program derived from $\Delta$) contains, for each $X \in A \cup \Sigma \cup \Pi$, a rule  
{\small
\[
X \leftarrow 
	\bigwedge_{\alpha \in \Sigma \wedge {\bf t}(\alpha)=X} (\neg \alpha \vee \neg \sb(\alpha))\wedge 
	 \bigwedge_{\beta \in \Pi \wedge \tb(\beta)=X} (\neg \beta \vee \sb(\beta)). 
\]
} 
\end{definition}

The set of complete extensions of an RAFN $\Delta$ coincides with the set of 
PSMs of $P_{\Delta}$.

\begin{theorem}\label{RAFN:compare-semantics}
	For any RAFN $\Delta$, \ \  	
$\widehat{{\cal CO}(\Delta)} = {\cal PS}(P_{\Delta})$.
\end{theorem}

Previous results also apply to restricted frameworks such as RAF, where $\Pi = \emptyset$, and AFN, where ${\bf t}:\Sigma \rightarrow A$.

\vspace*{1mm}
\noindent
{\bf ASAF.} We next provide definitions of defeated and acceptable sets for an ASAF.
\begin{definition}
Given an ASAF $\< A, \Sigma, \Pi, {\bf s}, {\bf t} \>$ and a set $\Set{S} \subseteq A \cup \Sigma \cup \Pi$, we define: 

\vspace*{1mm}
\noindent
$\bullet\ {\Def}(\Set{S}) = \{ X \in A \cup \Sigma \cup \Pi\ |\	
	{(X \in \Sigma \wedge \sb(X) \in \Def(\Sb))\  \vee}\ {(\exists \alpha \in \Sigma \cap \Set{S} \ldot 
	\tb(\alpha)=X)}\ \vee \\  
\hspace*{+46mm}	(\exists \beta \in \Pi \cap \Sb \ldot 
	\tb(\beta)=X\ \wedge\ \sb(\beta) \in \Def(\Sb))\}$; 
	
\vspace*{1mm}
\noindent
$\bullet\ \Acc(\Set{S})\! =\!\{\! X \!\in\! A\cup \Sigma\cup \Pi\ |\   (X \in \Sigma\! \imply \sb(X) \in\! \Acc(\Sb)) \wedge  (\forall \alpha \in \Sigma \ldot \ {\bf t}(\alpha) \= X \imply  \alpha \in \Def(\Sb)) \\
\hspace*{+40mm} \wedge (\forall \beta \in \Pi \ldot  \tb(\beta)\!=\!X\! \imply (\beta\!\in\! \Def(\Sb)~\vee~\sb(\beta)\!\in\! \Acc(\Set{S})))\}$.
\end{definition}

The 
acceptable elements of an ASAF can be computed by using the previous definition.

\begin{theorem}\label{AFRAS:Acceptable-Sets-Equivalence}
Given an ASAF $\Delta$ and an extension $\Set{S} \in {\cal CO}(\Delta)$, then $Def(\Sb) \= \Def(\Sb)$ and $Acc(\Sb) \= \Acc(\Sb)$.
\end{theorem}

By exploiting the result of Theorem~\ref{AFRAS:Acceptable-Sets-Equivalence}
now define the propositional program for an ASAF $\Delta$, which is easily derived by looking at the new definition of acceptable elements (i.e., $\Acc(\Sb)$).

\begin{definition}\label{def:AFRAS-program}
For any ASAF 
$\Delta = \<A, \Sigma, \Pi, {\bf s}, {\bf t}\>$, 
$P_\Delta $ (the propositional program derived from $\Delta$) contains, for each $X \in A \cup \Sigma \cup \Pi$, a rule of the form
\blue{\small
\[
X \leftarrow \varphi(X)
\wedge 
	\bigwedge_{\alpha \in \Sigma \wedge {\bf t}(\alpha)=X} \neg \alpha \wedge
	\bigwedge_{\beta \in \Pi \wedge \tb(\beta)=X} (\neg \beta \vee \sb(\beta)) 
	\ \text{where}\ 
	\varphi(X)=\begin{cases} {\bf s}(X)\ \text{if} \ X\in\Sigma\\
	\true\ \text{otherwise}.\\
	\end{cases}  
\]
}
\end{definition}
\begin{theorem}\label{ASAF:equivalence-psm}
	For any ASAF $\Delta$, \ \  	
$\widehat{{\cal CO}(\Delta)} = {\cal PS}(P_{\Delta} )$.
\end{theorem}

Similarly to the case of RAFN, the above results also apply to restricted frameworks such as AFRA, where $\Pi = \emptyset$, and AFN, where ${\bf t}:\Sigma \rightarrow A$.

\begin{example}\label{ex:RAFN-necessary-standard}\rm
Consider the Rec-BAF $\Delta'$ of Example \ref{ex:prel-afras} derived from the Rec-BAF of Figure~\ref{fig:asaf-two} by adding an argument $\tt s$ attacking argument $\tt w_t$ through $\alpha_4$, under the necessary interpretation of supports.
The propositional program under the RAFN semantics is $P_{\Delta'}=\{ \tt (w_i\leftarrow),$ 
$\tt (r\leftarrow \neg \alpha_1\vee \neg w_i),$ 
$\tt (w_e \leftarrow \neg \beta_1 \vee r),$ 
$\tt (p \leftarrow \neg \alpha_2 \vee \neg w_e),$ 
$\tt (w_t \leftarrow \neg \alpha_4 \vee \neg s),$ 
$\tt (\alpha_1 \leftarrow \neg \alpha_3 \vee \neg w_t),$ 
$\tt (s\leftarrow),$
$\tt (\alpha_2 \leftarrow ),$ 
$\tt (\alpha_3 \leftarrow ),$
$\tt (\alpha_4 \leftarrow ),$
$\tt  (\beta_1\leftarrow ) \}$, whose set of partial stable model is 
$M_1 = {\cal PS}(P_{\Delta'})=\{\{\tt s, w_i,\neg r, \neg w_e,\neg w_t, p,\beta_1, \alpha_1,\alpha_2, \alpha_3,\alpha_4\}\}$.

Analogously, the propositional program for $\Delta'$ under the ASAF semantics is $P_{\Delta'}=\{ \tt (w_i\leftarrow),$ 
$\tt (r\leftarrow \neg \alpha_1),$ 
$\tt (w_e \leftarrow \neg \beta_1 \vee r),$ 
$\tt (p \leftarrow \neg \alpha_2),$ 
$\tt (w_t \leftarrow \neg \alpha_4),$ 
$\tt (\alpha_1 \leftarrow \neg \alpha_3 \wedge w_i),$ 
$\tt (s\leftarrow),$
$\tt (\alpha_2 \leftarrow w_e),$ 
$\tt (\alpha_3 \leftarrow w_t),$
$\tt (\alpha_4 \leftarrow s),$
$\tt  (\beta_1\leftarrow ) \}$, whose set of partial stable model is $M_2 = {\cal PS}(P_{\Delta'})=\{\{\tt s, w_i,$ $\tt\neg r,\neg w_e,\neg w_t, p,\beta_1, \alpha_1,\neg \alpha_2, \neg \alpha_3,\alpha_4\}\}$, which differs from $M_1$ in the status of $\alpha_2$ and $\alpha_3$.~\hfill~$\square$
\end{example}

\section{Recursive BAFs with Deductive Supports}\label{sec:NovelSemanticsAFRS}
In this section we study two new frameworks both belonging to the Rec-BAF class and both extending AFD by allowing recursive attacks and deductive supports. 
The first one, called \emph{Recursive Argumentation Framework with Deductive supports (RAFD)}, extends RAF, whereas the second one, called \emph{Argumentation Framework with Recursive Attacks and Deductive supports (AFRAD)}, extends AFRA.  It is again assumed that $\Pi$ is acyclic and $\Sigma \cap \Pi = \emptyset$.

As we shall define the semantics by defining directly the sets $\Def(\Sb)$ and $\Acc(\Sb)$, differently from the previous section, we do not have any results regarding the equivalence between the sets $Acc(\Sb)$ and $\Acc(\Sb)$ for $\Sb = \Acc(\Sb)$.

\vspace*{1mm}
\noindent
{\bf RAFD.}
As usual, we first define the sets of defeated and acceptable elements, and then the propositional logic program for an RAFD.

\begin{definition} 
For any RAFD $\< A, \Sigma, \Pi, {\bf s}, {\bf t} \>$ and set $\Set{S} \subseteq A \cup \Sigma \cup \Pi$, we have that: 

\vspace*{1mm}
\noindent
$\bullet\ {\Def}(\Set{S})\! =\!\{ X \in A \cup \Sigma \cup \Pi\ |\ 
	(\exists \alpha \in \Sigma \cap \Set{S} \ldot  
	\tb(\alpha)=X \wedge \sb(\alpha) \in \Sb ) \ \vee \\ 
\hspace*{+46mm}
(\exists \beta \in \Pi \cap \Set{S} \ldot \sb(\beta)=X \wedge \tb~(\beta) \in \Def(\Sb) )\ \}$;

\vspace*{1mm}
\noindent
$\bullet\ \Acc(\Set{S})\! =\!\{ X \in A\cup \Sigma\cup \Pi\ |\ 
(\forall \alpha\! \in\! \Sigma \ldot \tb(\alpha) = X \imply (\alpha \in \Def(\Set{S}) \vee {\bf s}(\alpha) \in \Def(\Set{S})))\wedge 
\\ \hspace*{+46mm}  (\forall \beta\! \in\! \Pi\ldot \sb(\beta) = X\ \imply    (\beta \in \Def(\Set{S}) \vee \tb(\beta) \in \Acc(\Set{S})))\ \}$.
\end{definition}
The sets of extensions prescribed by the different semantics are based on the defeated and acceptable sets defined above.
That is, given an RAFD $\Delta=\< A, \Sigma, \Pi, {\bf s}, {\bf t} \>$, 
a set $\Set{S}\subseteq A\cup \Sigma\cup\Pi$ of elements is 
a complete extension of $\Delta$ \iff it is conflict-free (i.e., $\Set{S}\cap \Def(\Set{S}) = \emptyset$) and $\Set{S} = \Acc(\Set{S})$.
As done for the other frameworks, we use 
${\cal CO}(\Delta)$ to denote the set of complete extensions of $\Delta$.
Moreover, the set of preferred (resp., stable, semi-stable, grounded, ideal) extensions 
is defined in the standard way (see Section \ref{sec:AF}) 
by using again ${\Def}(\Set{S})$ and $\Acc(\Set{S})$.

Using the definition of $\Acc(\Sb)$, we define the propositional program for an RAFD $\Delta$.

\begin{definition}
Given an RAFD
$\Delta = \<A, \Sigma, \Pi, {\bf s}, {\bf t}\>$, then 
$P_\Delta $ (the propositional program derived from $\Delta$) contains, for each $X \in A \cup \Sigma \cup \Pi$, a rule of the form  
{\small
\[
X \leftarrow 
	\bigwedge_{\alpha \in \Sigma \wedge {\bf t}(\alpha)=X} (\neg \alpha \vee \neg \sb(\alpha))\wedge 
	 \bigwedge_{\beta \in \Pi \wedge \sb(\beta)=X} (\neg \beta \vee \tb(\beta)). 
\] 
}
\end{definition}

\begin{theorem}\label{RAFD:def}
	For any RAFD $\Delta$, \ \  	
$\widehat{{\cal CO}(\Delta)} = {\cal PS}(P_{\Delta} )$.
\end{theorem}

Thus, as expected, the semantics of an RAFD $\Delta$ can be carried out by using the PSMs of $P_\Delta$.

\vspace*{2mm}
\noindent
{\bf AFRAD.}
The following definition formalizes defeated and acceptable sets for an AFRAD.

\begin{definition}
Given an AFRAD $\< A, \Sigma, \Pi, {\bf s}, {\bf t} \>$ and a set $\Set{S} \subseteq A \cup \Sigma \cup \Pi$, we have that

\noindent
$\bullet\ {\Def}(\Set{S}) = \{ X \in A \cup \Sigma \cup \Pi\ |\	
	{(X \in \Sigma \wedge \sb(X) \in \Def(\Sb))\  \vee}\ {(\exists \alpha \in \Sigma \cap \Set{S} \ldot 
	\tb(\alpha)=X)}\ \vee \\  
\hspace*{+46.5mm}	(\exists \beta \in \Pi \cap \Sb \ldot 
	\sb(\beta)=X\ \wedge\ \tb(\beta) \in \Def(\Sb))\}$; 

\vspace*{1mm}
\noindent
$\bullet\Acc(\Set{S}) = \{\! X\! \in A\cup \Sigma\cup \Pi\ | \
	(X \!\in \!\Sigma\! \imply\! \sb(X) \in \Acc(\Sb))) \wedge (\forall \alpha \in \Sigma \ldot 
	{\bf t}(\alpha) \= X \imply\! \alpha \in \Def(\Sb)) \\
\hspace*{+45mm}
\wedge  (\forall \beta \in \Pi \ldot  \sb(\beta) = X\! \imply (\beta\!\in\! \Def(\Sb) \vee \tb(\beta) \!\in\! \Acc(\Sb))))\}$.
\end{definition}

Similarly to what done for RAFDs, 
the set ${\cal CO}(\Delta)$ of complete extensions of an AFRAD $\Delta$, 
and the sets of extensions prescribed by the other semantics, are 
defined by using the defeated and acceptable sets defined above.

\begin{definition}\label{def:AFRAS-program-deductive}
For any AFRAD
$\Delta = \<A, \Sigma, \Pi, {\bf s}, {\bf t}\>$, 
$P_\Delta $ (the propositional program derived from $\Delta$) contains, for each $X \in A \cup \Sigma \cup \Pi$, a rule of the form
\blue{\small
\[
X \leftarrow 
	\varphi(X) \wedge 
	\bigwedge_{\alpha \in \Sigma \wedge {\bf t}(\alpha)=X} \neg \alpha  \wedge
	 \bigwedge_{\beta \in \Pi \wedge \sb(\beta)=X} (\neg \beta \vee \tb(\beta))
	  \ \text{where}\ 
	\varphi(X)=\begin{cases} {\bf s}(X)\ \text{if} \ X\in\Sigma\\
	\true\ \text{otherwise}.\\
	\end{cases}  
\]
}

\end{definition}

\begin{theorem}\label{AFRA:def}
	For any AFRAD $\Delta$, \ \  	
$\widehat{{\cal CO}(\Delta)} = {\cal PS}(P_{\Delta} )$.
\end{theorem}

\begin{example}\label{ex:RAFN-necessary-standard-cont}\rm
Consider the Rec-BAF $\Delta'$ of Example \ref{ex:prel-afras}
and assume that supports are interpreted as deductive.
The propositional program under the  RAFD semantics
is $P_{\Delta'}=\{ \tt (w_i\leftarrow),$ 
$\tt (r\leftarrow (\neg \alpha_1\vee \neg w_i) \wedge (\neg \beta_1 \vee w_e)),$ 
$\tt (w_e \leftarrow ),$
$\tt (p \leftarrow \neg \alpha_2 \vee \neg w_e),$ 
$\tt (w_t \leftarrow \neg \alpha_4 \vee \neg s),$ 
$\tt (s\leftarrow),$
$\tt (\alpha_1 \leftarrow \neg \alpha_3 \vee \neg w_t),$ 
$\tt (\alpha_2 \leftarrow ),$ 
$\tt (\alpha_3 \leftarrow ),$
$\tt (\alpha_4 \leftarrow ),$
$\tt  (\beta_1\leftarrow ) \}$, whose set of partial stable model is $M_1 = {\cal PS}(P_{\Delta'})=\{\{\tt s, w_i,\neg r,  w_e,\neg w_t, \neg p,\beta_1, \alpha_1,\alpha_2, \alpha_3,\alpha_4\}\}$. Analogously, the propositional program for $\Delta'$ under the AFRAD semantics is $P_{\Delta'}=\{ \tt (w_i\leftarrow),$ 
{$\tt (r\leftarrow \neg \alpha_1 \wedge (\neg \beta_1 \vee w_e)),$}
{$\tt (w_e \leftarrow ),$}
$\tt (p \leftarrow \neg \alpha_2),$ 
$\tt (w_t \leftarrow \neg \alpha_4),$ 
$\tt (s\leftarrow),$
$\tt (\alpha_1 \leftarrow  w_i\wedge \neg \alpha_3),$ 
$\tt (\alpha_2 \leftarrow w_e),$ 
$\tt (\alpha_3 \leftarrow w_t),$
$\tt (\alpha_4 \leftarrow s),$
$\tt  (\beta_1\leftarrow ) \}$, whose set of partial stable model is 
$M_2 = {\cal PS}(P_{\Delta'})=\{\{\tt s, w_i,\neg r, w_e,\neg w_t, \neg p, \alpha_1,\neg \alpha_2, \neg \alpha_3,\alpha_4,\beta_1\}\}$. Observe that the RAFD (resp., AFRAD) program differs from the RAFN (resp., ASAF) program only in rules having as head arguments $\tt r$ and $\tt w_e$.~\hfill~$\square$
\end{example}


\section{Discussion and Future Work}\label{sec:conclusions}
By exploring the connection between formal argumentation and logic programming, we have proposed a simple but general logical framework  which is able to capture, in a systematic and succinct way, the different features of several AF-based frameworks under different argumentation semantics and  interpretation of the support relation.
The proposed approach can be used for better understanding  the semantics of extended AF frameworks (sometimes a bit involved), and is flexible enough for encouraging the study of other extensions.

As pointed out in Section~\ref{Sec_Intro},
our work is complementary to approaches
providing the semantics for an AF-based framework
by using meta-argumentation,
that is, by relying on a translation from a given AF-based framework 
to an AF~\cite{CohenGGS15}.
In this regard, 
we observe that meta-argumentation approaches have 
the drawback of making a bit difficult understanding the original 
meaning of arguments and interactions once translated into the resulting meta-AF.
In fact, those approaches rely on translations that generally require 
adding several meta-arguments 
and meta-attacks to the resulting meta-AF in order to model 
the original interactions.

Concerning approaches that provide
the semantics of argumentation frameworks by LPs~\cite{CaminadaSAD15},
we observe that
a logic program for an AF-based framework can be obtained 
by first flattening the given framework into a meta-AF 
and then converting it into a logic program.
The so-obtained program contains the translation 
of meta-arguments and meta-attacks that
make the program much more verbose 
and difficult to understand 
(because not straightly derived from the given extended AF framework)
in our opinion,
compared with the direct translation we proposed.
For instance, 
given an ASAF $\Delta = \<A, \Sigma, \Pi, {\bf s}, {\bf t}\>$,
the propositional program $P_{\Delta}$ directly obtained from $\Delta$
has a number of rules equal to $|A|+ |\Sigma|+ |\Pi|$,
while the program $P_{\Delta'}$ obtained 
considering
the translation from $\Delta$ to an AFN and then to  meta-AF $\Delta'$
consists of $|A|+|\Sigma|+3|\Pi|$ rules,
of which $2|\Pi|$ rules define new (meta-)arguments
(examples of the arguments introduced, 
which correspond to rules of $P_{\Delta'}$,
can be found in~\cite{Gottifredi-Cohen-Garcia-Simari18}).
In addition, 
the size of body's rules may also increase for $P_{\Delta'}$
since the number of extended/complex attacks
that need to be added may be relevant in some cases.
Finally, the models of $P_{\Delta'}$ 
contain literals corresponding to meta-arguments  having no 
meaning w.r.t. extensions of the extended AF $\Delta$.

In brief, the program that we directly obtain 
from a given AF-based framework
is more concise and easy to understand with respect to that 
obtained by (possibly several stages of) translations to AF.
Moreover, the proposed approach 
uniformly deals with several AF-based frameworks,
including 
RAFN and the novel frameworks RAFD and AFRAD
for which a translation to AF has not been defined.
Nevertheless, we believe that our approach is also
complementary to approaches using intermediate translations to AF in order to define an LP for an extended AF.

Furthermore, 
our approach
can also be used to provide additional tools for computing complete extensions  using answer set solvers \cite{ASP-Systems} and classical program rewriting \cite{unfolding,SakamaR17,GagglMRWW15}. 
\blue{In particular, 
we plan to experimentally compare 
the following LP approaches for 
the computation of extensions of AF-based frameworks $\Delta$:
(i) 
using the propositional program $P_{\Delta}$ directly obtained from $\Delta$;
and 
(ii) 
using the propositional program $P_{\Delta'}$ obtained from $\Delta$ by transforming it to an AF $\Delta'$ (possibly through different transformations, involving different intermediate argumentation frameworks).
}

Other extensions of the Dung's framework not explicitly discussed in this paper are also captured by our technique as they are special cases of some of those studied in this paper.
This is the case of \emph{Extended AF (EAF)} and \emph{hierarchical EAF}, which extend AF by allowing second order and stratified attacks, respectively \cite{Modgil-AI-09}, that are special cases of recursive attacks.

Future work will be also devoted to further generalize our logical approach in order to deal also with AF-based framework considering probabilities~\cite{FazzingaFP15},
weights~\cite{BistarelliRS18},
and preferences~\cite{AmgoudV11,Modgil-AI-09}, and frameworks considering supports with multiple sources~\cite{CayrolFCL18}.
Finally, we plan to investigate incremental techniques tailored 
at using our approach 
to compute extensions of dynamic AF-based frameworks, 
where the sets of arguments and interactions change over the time~\cite{GrecoP16,AlfanoGP17,AlfanoGP18,ecai2020}.

\bibliographystyle{acmtrans}
\bibliography{ICLP-TPLP-CR}
\label{lastpage}

\newpage
\section*{Appendix A: Proofs}
\setcounter{theorem}{0}
In this appendix, the interested reader can find  the proofs of the results stated in the paper.

\vspace*{3mm}
\noindent
\textit{Theorem \ref{BAFn:Acceptable-Sets-Equivalence}.\\
Given an AFN $\Delta$ and an extension $\Set{S} \in {\cal CO}(\Delta)$, then $Def(\Sb) \= \Def(\Sb)$ and $Acc(\Sb) \= \Acc(\Sb)$. }

\vspace*{2mm}
\noindent
\textit{Proof.}
We prove the theorem by introducing a lemma showing that, for AFNs, mediated attacks do not affect the status of arguments when $\Sb$ is a complete extension.

\begin{lemma}\label{Lemma1}
For any AFN $\Delta = \<A,\Omega, \Gamma\>$, 
let $\Delta_n$ be the AF derived from $\Delta$ and $\Delta'_n$ the AF derived from $\Delta_n$ by deleting mediated attacks, 
then ${\cal CO}(\Delta) = {\cal CO}(\Delta_n) = {\cal CO}(\Delta'_n)$.
\end{lemma}
\noindent 
\emph{Proof.}
Consider a mediated attack $a \rightarrow b$ in $\Delta_n$ derived from a (possibly transitive) support $c \supportplus a$ and an attack $c \rightarrow b$ in $\Delta$. 
The status of argument $b$ w.r.t. a complete extension $\Sb$ is not influenced by the status of $a$, but it is determined by other arguments attacking or supporting it. 
This is carried out by considering the possible status of $c$:
\begin{itemize}
\item
$c \in \Sb$: this means that $b \in Def(\Sb)$, independently from the status of $a$, as $c$ is attacking directly $b$.
\item
$c \in Def(\Sb)$: this means that also $a \in Def(\Sb)$ and, therefore, the attacks of both arguments $c$ and $a$ are not relevant for the status of $b$.
\item 
$c \not\in \Sb \cup Def(\Sb)$: if $c$ is undecided, then $a \not\in \Sb$, that is either $a \in Def(\Sb)$ (i.e. it is false) or $c \not\in \Sb \cup Def(\Sb)$ (i.e. it is undecided). 
In both cases $a$ is not relevant to determine the status of $b$. Indeed, i) if $a \in Def(\Sb)$ the mediated attack from $a$ to $b$ is not relevant, whereas ii) if $a$ is undecided, since $c$ is also undecided, eliminating the mediated attack from $a$ to $b$ does not change the status of $b$. 
\hfill $\Box$
\end{itemize}

\vspace*{2mm}
We now show the equivalence of $Def(\Sb)$ and $\Def(\Sb)$, that is, for any AFN $\Delta = \<A,\Omega, \Gamma\>$, and set $\Sb \in {\cal CO}(\Delta)$, $Def(\Sb) = \Def(\Sb)$.

Let $\Delta_n = \<A,\Omega_n\>$ be the AF derived from $\Delta$, we have that
$Def(\Sb)\= \{ a \in A\! \mid\! \exists b \in \Sb . (b,a) \in \Omega_n \}$.
This set can be rewritten as 
$Def(\Sb) = \{ a \in A \mid \exists b \in \Sb . (b,a) \in \Omega \vee \exists c \in A. (b,c) \in \Omega \wedge (c,a) \in \Gamma^+  \}$ 
which is equivalent to 
$Def(\Sb) = \{ a \in A \mid \exists b \in \Sb . (b,a) \in \Omega \vee \exists c \in Def(\Sb). (c,a) \in \Gamma^+  \}$.
Moreover, if we have a sequence of supports $c_1 \Rightarrow \cdots \Rightarrow c_n$ with $c_1 \in Def(\Sb)$, then $c_i \in Def(\Sb)$ for all $i \in [1,n]$.
This implies that we can rewrite the set of defeated elements as
$Def(\Sb) = \{ a \in A \mid \exists b \in \Sb . (b,a) \in \Omega \vee \exists c \in Def(\Sb). (c,a) \in \Gamma \}$ which is equal to $\Def(\Sb).$

\vspace*{2mm}
We now continue with the proof of the theorem and show the equivalence of $Acc(\Sb)$ and $\Acc(\Sb)$, that is, for any AFN $\Delta = \<A,\Omega, \Gamma\>$, and set $\Sb \in {\cal CO}(\Delta)$, $Acc(\Sb) = \Acc(\Sb)$.

\noindent 
Let $\Delta_n = \<A,\Omega_n\>$ be the AF derived from $\Delta$, 
$Acc(\Set{S})\, = \{a \in A\ \mid\ \forall b \in A\ldot(b,a) \in \Omega_n\imply~b \in Def(\Sb)\}$. 
Then, 
$Acc(\Sb) = \{ a \in A \mid (\forall b \in A . (b,a) \in \Omega \imply b \in Def(\Sb)) \wedge (\forall c \in A. (b,c) \in \Omega \wedge (c,a) \in \Gamma^+ \imply b \in Def(\Sb) )\}$ 
which can be rewritten as 
$Acc(\Sb) = \{ a \in A \mid (\forall b \in A . (b,a) \in \Omega \imply b \in Def(\Sb))\wedge (\forall c \in A. (c,a) \in \Gamma^+ \imply c \in Def(\Sb)) \}$ 
and is equivalent to 
$Acc(\Sb) = \{ a \in A \mid (\forall b \in A . (b,a) \in \Omega \imply b \in Def(\Sb)) \wedge (\forall c \in A. (c,a) \in \Gamma \imply c \in Def(\Sb)\}$ which is equal to $\Acc(\Sb).$ 
\hfill $\Box$

\vspace*{3mm}

\newpage
\noindent
\textit{Theorem \ref{BAFn:semantics-equivalence}.\\
For any AFN $\Delta$, \ \ $\widehat{{\cal CO}(\Delta)} = {\cal PS}(P_{\Delta} ) = {\cal PS}(P_{\Deltan})$.}

\noindent 
\emph{Proof.} 
$\widehat{{\cal CO}(\Delta)} = {\cal PS}(P_{\Deltan})$ derives from the fact that $\widehat{{\cal CO}(\Delta)} = \widehat{{\cal CO}(\Delta_n)}$ and  $\widehat{{\cal CO}(\Delta_n)} = {\cal PS}(P_{\Deltan})$.
We prove now that $\widehat{{\cal CO}(\Delta)} = {\cal PS}(P_{\Delta} )$.
\begin{itemize}
\item
$\widehat{{\cal CO}(\Delta)} \subseteq {\cal PS}(P_{\Delta}).$
We prove that for any $\Sb \in {\cal CO}(\Delta)$, $\widehat{\Sb} \in {\cal PS}(P_{\Delta})$.
Indeed, $P_\Delta$ contains, for each atom $a \in A$, a rule 
$a \leftarrow \bigwedge_{(b,a) \in \Omega} \neg b \wedge \bigwedge_{(c,a) \in \Gamma} c$.
Moreover, $P_\Delta^{\widehat{\Sb}}$ (the positive instantiation of $P_\Delta$ w.r.t. $\widehat{\Sb}$) contains positive rules defining exactly the arguments in $\Sb$, whose bodies contains only (positive) arguments in $\Sb$.
Since the relation $\Gamma$ is acyclic we have that 
$T_{P_\Delta^{\widehat{\Sb}}}^\omega(\emptyset) = \Sb$, that is $\Sb$ is a PSM for $P_\Delta^{\widehat{\Sb}}$.
\item
${\cal PS}(P_{\Delta}) \subseteq \widehat{{\cal CO}(\Delta)}.$
Consider a PSM $M \in {\cal PS}(P_\Delta)$, $pos(M) = T^\omega_{P_\Delta^M}(\emptyset)$. \\
$pos(M) \subseteq A$ is conflict free w.r.t. $\Delta$. Indeed, assuming that there are two arguments $a,b \in pos(M)$ such that $(a,b) \in \Omega$, this means that the rule defining $b$ in $P_\Delta$ contains in the body a literal $\neg a$. This is not possible as in such a case $b \not\in T^\omega_{P_\Delta^M}(\emptyset)$.
Assuming that $a$ attacks $b$ indirectly through a supported attack $a \rightarrow a_1 \Rightarrow \cdots \Rightarrow a_n \Rightarrow b$. In such a case $a_1,...,a_n,b \not\in T^\omega_{P_\Delta^M}(\emptyset)$. Thus, $pos(M)$ is conflict free.\\
Moreover, from Definition \ref{ACC:def} and Theorem \ref{BAFn:Acceptable-Sets-Equivalence}, considering that $pos(M) = T^\omega_{P_\Delta^M}(\emptyset)$, we derive that $pos(M) = Acc(pos(M))$. 
\hfill $\Box$
\end{itemize}

\vspace*{3mm}
\noindent
\emph{Theorem \ref{BAFd:Acceptable-Sets-Equivalence}.\\
Given an AFD $\Delta$ and an extension $\Set{S} \in {\cal CO}(\Delta)$, then $Def(\Sb) \= \Def(\Sb)$ and $Acc(\Sb) \= \Acc(\Sb)$.}

\noindent 
\emph{Proof.}
The proof follows the one of Theorem \ref{BAFn:Acceptable-Sets-Equivalence}.
Whenever in the proof of Theorem \ref{BAFn:Acceptable-Sets-Equivalence}
we have a support $c \Rightarrow a$, here we have a support $a \Rightarrow c$.
As a consequence, while in Theorem \ref{BAFn:Acceptable-Sets-Equivalence} we discard mediated attacks (Lemma \ref{Lemma1}), here we discard supported attacks.
\hfill $\Box$

\vspace*{3mm}
\noindent
\emph{Theorem \ref{BAFd:semantics-equivalence}\\
For any AFD  $\Delta$, $\widehat{{\cal CO}(\Delta)} = {\cal PS}(P_{\Delta} ) = {\cal PS}(P_{\Deltad})$. }

\noindent
\emph{(Proof.)}
The proof is analogous to that of Theorem~\ref{BAFn:semantics-equivalence}.
\hfill $\Box$

\vspace*{3mm}
\noindent
\emph{Theorem \ref{RAFN:Acceptable-Sets-Equivalence}\\
Given an RAFN $\Delta$ and an extension $\Set{S} \in {\cal CO}(\Delta)$, then $Def(\Sb) \= \Def(\Sb)$ and $Acc(\Sb) \= \Acc(\Sb)$. }

\vspace*{2mm}
\noindent 
\emph{(Proof.)}
We first prove the equivalence of $Def(\Sb)$ and $\Def(\Sb)$, that is, for any RAFN $\Delta = \<A, \Sigma, \Pi, {\bf s}, {\bf t}\>$, and set $\Sb \in {\cal CO}(\Delta)$, $Def(\Sb) = \Def(\Sb)$.
Recalling that 
$Def(\Sb) = \{ X \in A \cup \Sigma \cup \Pi\ |\ 
	\exists b \in A \cap \Sb \ldot 
	b\ {\tt att_\Sb}\ X\}$, and 
using the definition of $b\ {\tt att_\Sb}\ X$, we can rewrite $Def(\Sb)$ as follows: \\
$Def(S) = \ \{ X \in A \cup \Sigma \cup \Pi\ |\ 
	\exists \alpha \in \Sigma \cap \Sb \ldot 
	{\bf s}(\alpha) \in \Sb \wedge (\tb(\alpha) = X \vee (\tb(\alpha),X) \in (\Pi \cap \Sb)^+) \}$.

Moreover, when $\tb(\alpha) \in Def(\Sb)$, all arguments involved in the supported attack from $\sb(\alpha)$ to $X$ are defeated. Thus, $Def(\Sb)$ ca be rewritten as follows: \\
$Def(\Sb) =  \{ X \in A \cup \Sigma \cup \Pi\ |\ 
	(\exists \alpha \in \Sigma \cap \Sb \ldot 
	\sb(\alpha) \in \Sb \wedge \tb(\alpha) = X) \vee \ (\exists \beta \in \Pi \cap \Sb \ldot \sb(\beta) \in Def(\Sb) \wedge \tb(\beta) = X) \}$.
Therefore, $Def(\Sb) = \Def(\Sb)$.

\vspace*{2mm}
We now continue with the proof of the theorem.
To this end we perform analogous transformations to those used above and in Theorem \ref{BAFn:Acceptable-Sets-Equivalence}.

Recalling that 
$Acc(\Set{S})\, = \{ X \in A \cup \Sigma \cup \Pi\ |\ 
	\forall b \in A \ldot b\ {\tt att_\Sb}\ X \imply b \in  Def(\Sb)\}$ and using the definition of $b\ {\tt att_\Sb}\ X$, the set of acceptable elements can be rewritten as follows:
$Acc(\Set{S})\, = \{ X \in A \cup \Sigma \cup \Pi\ |\ 
	\forall \alpha \in \Sigma \ldot \sb(\alpha)\ {\tt att_\Sb}\ X \imply \sb(\alpha) \in  Def(\Sb)\}$ which is equivalent to 
$Acc(\Set{S})\, = \{ X \in A \cup \Sigma \cup \Pi\ |\ 
	\forall \alpha \in \Sigma \cap \Sb \ldot \sb(\alpha)\ {\tt att_\Sb}\ X \imply \sb(\alpha) \in  Def(\Sb)\}$.
		
Moreover,  $\forall \alpha \in \Sigma \cap \Sb \ldot \sb(\alpha)\ {\tt att_\Sb}\ X \imply \sb(\alpha) \in  Def(\Sb)$ is equivalent to the formula
$\forall \alpha \in \Sigma \cap \Sb\ \forall \beta_1,...,\beta_n \in \Pi \cap \Sb \ldot  \tb(\alpha) = \sb(\beta_1) \wedge \bigwedge_{i\in[1,{n})} (\tb(\beta_i) = \sb(\beta_{i+1})) \wedge \tb(\beta_n) = X \imply \sb(\alpha) \in  Def(\Sb)$.

Under the condition $\forall \alpha \in \Sigma \cap \Sb\ \forall \beta_1,...,\beta_n \in \Pi \cap \Sb \ldot  \tb(\alpha) = \sb(\beta_1) \wedge \bigwedge_{i\in[1,{n})} (\tb(\beta_i) = \sb(\beta_{i+1})) \wedge \tb(\beta_n) = X$, we have that $\sb(\alpha) \in  Def(\Sb)$ iff $\sb(\beta_n) \in \Sb$ (which is equal to $Acc(\Sb))$.
Therefore, the set of acceptable elements can be rewritten as follows:
$Acc(\Set{S})\, = \{ X \in A \cup \Sigma \cup \Pi\ |\ 
	(\forall \alpha \in \Sigma \cap \Sb \ldot \tb(\alpha) = X \imply \sb(\alpha) \in  Def(\Sb)) \wedge (\forall \beta \in \Pi \cap \Sb \ldot \tb(\beta) = X \imply \sb({\beta}) \in \Sb) \}$.
	
The last formula can equivalently be rewritten (moving $\alpha \in \Sb$ and $\beta \in \Sb$ from the bodies to the heads of the two implications) as:
$Acc(\Set{S})\, = \{ X \in A \cup \Sigma \cup \Pi\ |\ 
	(\forall \alpha \in \Sigma \ldot \tb(\alpha) = X \imply \sb(\alpha) \in  Def(\Sb) \vee \alpha \in Def(\Sb)) \ \wedge\ (\forall \beta \in \Pi \ldot \tb(\beta) = X \imply \sb(\beta) \in \Sb \vee \beta \in Def(\Sb)) \}$.
Thus, recalling that for complete extensions $\Sb = Acc(\Sb)$, we have proved that $Acc(\Sb) = \Acc(\Sb)$.
\hfill $\Box$

\vspace*{3mm}

\vspace*{3mm}
\noindent
\emph{Theorem \ref{RAFN:compare-semantics}\\
	For any RAFN $\Delta$, \ \  	
$\widehat{{\cal CO}(\Delta)} = {\cal PS}(P_{\Delta})$. }

\noindent
\emph{(Proof.)}
The proof follows the one of Theorem \ref{BAFn:semantics-equivalence}. The only difference is that complete extensions also contain attacks and supports, whereas the logic program also contains rules defining attacks and supports and, consequently, the partial stable models contain arguments, attacks and supports.
\hfill $\Box$

\vspace*{3mm}
\noindent
\emph{Theorem \ref{AFRAS:Acceptable-Sets-Equivalence}\\
Given an ASAF $\Delta$ and an extension $\Set{S} \in {\cal CO}(\Delta)$, then $Def(\Sb) \= \Def(\Sb)$ and $Acc(\Sb) \= \Acc(\Sb)$. }

\vspace*{2mm}
\noindent
\emph{(Proof.)}
We first prove the equivalence of $Def(\Sb)$ and $\Def(\Sb)$, that is, for any ASAF $\Delta = \<A,\Sigma,\Pi,\sb,\tb\>$, and set $\Sb \in {\cal CO}(\Delta)$, $Def(\Sb) = \Def(\Sb)$.

Recalling that 
$Def(\Sb) = \{ X \in A \cup \Sigma \cup \Pi\ |\ 
	\exists b \in A \cap \Sb \ldot 
	b\ {\tt def_\Sb}\ X\}$, and 
using the definition of $b\ {\tt def_\Sb}\ X$, we can rewrite $Def(\Sb)$ as follows: \\
$Def(\Sb) = \ \{ X \in A \cup \Sigma \cup \Pi\ |\ 
	(X \in \Sigma \wedge \sb(X) \in Def(\Sb)) \vee 
	(\exists \alpha \in \Sigma \cap \Sb \ldot 
	\tb(\alpha) = X \vee (\tb(\alpha),X) \in (\Pi \cap \Sb)^+) \}$.

Moreover, when $\tb(\alpha) \in Def(\Sb)$, all arguments involved in the extended  {defeat} from $\sb(\alpha)$ to $X$ are defeated. Thus, $Def(\Sb)$ ca be rewritten as follows: \\
$Def(\Sb) =  \{ X \in A \cup \Sigma \cup \Pi\ |\ 
	(X \in \Sigma \wedge \sb(X) \in Def(\Sb)) \vee 
	(\exists \alpha \in \Sigma \cap \Sb \ldot 
	\tb(\alpha) = X) \vee \ 
	(\exists \beta \in \Pi \cap \Sb \ldot \sb(\beta) \in Def(\Sb) \wedge \tb(\beta) = X \}$.
Therefore, $Def(\Sb) = \Def(\Sb)$.
	 
\vspace*{2mm}
We now continue with the proof of the theorem.

Recalling that 
$Acc(\Set{S})\, = \{ X \in A \cup \Sigma \cup \Pi\ |\ 
	\forall \alpha \in \Sigma \ldot \alpha\ {\tt def_\Sb}\ X \imply \alpha \in  Def(\Sb)\}$ and using the definition of $b\ {\tt def_\Sb}\ X$, the set of acceptable elements can be rewritten as follows:
$Acc(\Set{S})\, = \{ X \in A \cup \Sigma \cup \Pi\ |\ 
	\forall \alpha \in \Sigma \ldot \alpha\ {\tt def_\Sb}\ X \imply \alpha \in  Def(\Sb)\}$ which is equivalent to 
$Acc(\Set{S})\, = \{ X \in A \cup \Sigma \cup \Pi\ |\ 
	\forall \alpha \in \Sigma \cap \Sb \ldot \alpha\ {\tt def_\Sb}\ X \imply \alpha \in  Def(\Sb)\}$.
	
Moreover,  $\forall \alpha \in \Sigma \cap \Sb \ldot \alpha\ {\tt def_\Sb}\ X \imply \alpha \in  Def(\Sb)$ is equivalent to the formula
$\forall \alpha \in \Sigma\ \forall \beta_1,...,\beta_n \in \Pi \cap \Sb \ldot  \tb(\alpha) = \sb(\beta_1) \wedge \bigwedge_{i\in[1,{n})} (\tb(\beta_i) = \sb(\beta_{i+1})) \wedge \tb(\beta_n) = X \imply \alpha \in  Def(\Sb) \wedge (X \in \Sigma \imply \sb(\alpha) \in Def(\Sb))$.

Under the condition $\forall \alpha \in \Sigma\ \forall \beta_1,...,\beta_n \in \Pi \cap \Sb \ldot  \tb(\alpha) = \sb(\beta_1) \wedge \bigwedge_{i\in[1,{n})} (\tb(\beta_i) = \sb(\beta_{i+1})) \wedge \tb(\beta_n) = X$, we have that $\alpha \in  Def(\Sb)$ iff $\sb(\beta_n) \in \Sb$ (which is equal to $Acc(\Sb))$.
Therefore, the set of acceptable elements can be rewritten as follows:
$Acc(\Set{S})\, = \{ X \in A \cup \Sigma \cup \Pi\ |\ 
	(\forall \alpha \in \Sigma \ldot \tb(\alpha) = X \imply \alpha \in  Def(\Sb)) \wedge 
	(\forall \beta \in \Pi \cap \Sb \ldot \tb(\beta) = X \imply \sb({\beta}) \in \Sb)) \wedge 
	(X \in \Sigma \imply \sb(X) \in \Acc(\Sb) ) \}$.
	
The last formula can  equivalently be rewritten as:
$Acc(\Set{S})\, = \{ X \in A \cup \Sigma \cup \Pi\ |\ 
	(\forall \alpha \in \Sigma \ldot \tb(\alpha) = X \imply \alpha \in Def(\Sb)) \ \wedge\ (\forall \beta \in \Pi \ldot \tb(\beta) = X \imply \sb(\beta) \in \Sb \vee \beta \in Def(\Sb))  \wedge 
	(X \in \Sigma \imply \sb(X) \in \Acc(\Sb) ) \}$.
	Thus, recalling that for complete extensions $\Sb = Acc(\Sb)$, we have proved that $Acc(\Sb) = \Acc(\Sb)$.~\hfill~$\Box$

\vspace*{3mm}
\noindent
\emph{Theorem \ref{ASAF:equivalence-psm}\\
For any ASAF $\Delta$, \ \  	
$\widehat{{\cal CO}(\Delta)} = {\cal PS}(P_{\Delta} )$. }

\noindent
\emph{(Proof.)}
The proof follows that of Theorem~\ref{RAFN:compare-semantics}.
\hfill $\Box$

\vspace*{3mm}
\noindent
\emph{Theorem \ref{RAFD:def}\\
	For any RAFD $\Delta$, \ \  	
$\widehat{{\cal CO}(\Delta)} = {\cal PS}(P_{\Delta} )$. }

\noindent
\emph{(Proof.)}
The proof follows that of Theorem~\ref{RAFN:compare-semantics}.
\hfill $\Box$

\vspace*{3mm}
\noindent
\emph{Theorem \ref{AFRA:def}\\
	For any AFRAD $\Delta$, \ \  	
$\widehat{{\cal CO}(\Delta)} = {\cal PS}(P_{\Delta} )$. }

\noindent
\emph{(Proof.)}
The proof follows that of Theorem~\ref{RAFN:compare-semantics}.
\hfill $\Box$

\end{document}